\documentclass[manuscript, nonacm]{acmart} 

\usepackage{subfigure}
\usepackage{tabularx}
\usepackage[bb=dsserif]{mathalpha}

\AtBeginDocument{%
  \providecommand\BibTeX{{%
    \normalfont B\kern-0.5em{\scshape i\kern-0.25em b}\kern-0.8em\TeX}}}

\setcopyright{acmcopyright}
\copyrightyear{2023}
\acmYear{2023}
\acmDOI{XXXXXXX.XXXXXXX}

\acmConference[FAccT '23]{}{June 12--15,
  2023}{Chicago, IL}
\acmPrice{15.00}
\acmISBN{978-1-4503-XXXX-X/18/06}

\begin{document}

\title{On the Importance of Application-Grounded Experimental Design for Evaluating Explainable Machine Learning Methods}

\author{\NoCaseChange{Kasun Amarasinghe$^{1}$, Kit T. Rodolfa$^{1}$, S\'ergio Jesus$^{2}$, Valerie Chen$^{1}$,
Vladimir Balayan$^{2}$, Pedro Saleiro$^{2}$, Pedro Bizarro$^{2}$, Ameet Talwalkar$^{1}$, Rayid Ghani$^{1}$}}
 \affiliation{
    \institution{\\Carnegie Mellon University}
    \city{Pittsburgh}
    \state{Pennsylvania}
    \country{USA$^{1}$}
}
 \affiliation{
    \institution{Feedzai}
    \city{Lisbon}
    \country{Portugal$^{2}$}    
}




\begin{abstract}
Most existing evaluations of explainable machine learning (ML) methods rely on simplifying assumptions or proxies that do not reflect real-world use cases; the handful of more robust evaluations on real-world settings have shortcomings in their design, resulting in limited conclusions of methods' real-world utility.  
In this work, we seek to bridge this gap by conducting a study that evaluates three popular explainable ML methods in a setting consistent with the intended deployment context. 
We build on a previous study on e-commerce fraud detection and make crucial modifications to its setup relaxing the simplifying assumptions made in the original work that departed from the deployment context.
In doing so, we draw drastically different conclusions from the earlier work and find no evidence for the incremental utility of the tested methods in the task.
Our results highlight how seemingly trivial experimental design choices can yield misleading conclusions, with lessons about the necessity of not only evaluating explainable ML methods using tasks, data, users, and metrics grounded in the intended deployment contexts but also developing methods tailored to specific applications.
In addition, we believe the design of this experiment can serve as a template for future study designs evaluating explainable ML methods in other real-world contexts.  
\end{abstract}

\maketitle

\section{Introduction}

Despite the rapid expansion in method development for explainable machine learning (ML), research on metrics and methods for evaluating those methods' real-world utility has lagged \cite{amarasinghe2021explainable, chen2021interpretable}. 
Evaluation studies frequently rely on simplified experimental settings with non-expert users (e.g., workers on Amazon Mechanical Turk), use proxy tasks (e.g., forward simulation), or use subjective, user-reported measures as metrics of explanation quality \cite{islam2020towards,yalcin2021evaluating, ribeiro2016lime, ribeiro2018anchors, lundberg2018treeshap, lundberg2017unified, lakkaraju2016interpretable}. 
Such settings are not equipped to evaluate the real-world utility of explainable ML methods since proxy task performance does not reflect real-task performance \cite{buccina2020proxy}, users' perception of explanation usefulness is not reflective of utility in a task \cite{buccina2020proxy, lakkaraju2020how}, and proxy users do not reflect how expert users would use explanations \cite{amarasinghe2021explainable}.  
A few studies evaluate explainable ML methods on their intended deployment settings where domain expert users perform the intended task  \cite{lundberg2018explainable, jesus2021} (dubbed application-grounded evaluation studies in \cite{doshi2017towards}).
However, even in those, we argue that experimental design flaws (e.g., not isolating the incremental impact of explanations in \cite{lundberg2018explainable}) and seemingly trivial design choices that deviate experimental settings from the deployment context (e.g., using metrics that do not reflect the task objectives in \cite{jesus2021}), limit the applicability of drawn conclusions. We elaborate on these limitations in Section~\ref{sec: background}.   

In this work,  we seek to bridge this critical gap by conducting a study that evaluates explainable ML methods in a setting \textit{consistent with the intended deployment context}. 
Our study builds on the e-commerce fraud detection setting used in a previous evaluation study~\cite{jesus2021} consisting of professional fraud analysts tasked with reviewing e-commerce transactions to detect fraud when the ML model is uncertain about the outcome. 
We identify several simplifying assumptions made by the previous study that deviated from the deployment context and modify the setup to relax those assumptions (summarized in Table~\ref{tab:study-diffs} and discussed in detail in Section \ref{sec: modifications}). 
These modifications make the experimental setup faithful to the deployment setting and equipped to evaluate the utility of the explainable ML methods considered.  
Our setup results in \textit{dramatically different conclusions} of the relative utility of ML model scores and explanations compared to the earlier work \cite{jesus2021}. For instance, while the authors of \cite{jesus2021} find that the ML model scores and explanations impact confusion matrix-based metrics, we do not find any incremental utility of the ML models or explanations in improving the decision correctness metric that captures the operational objectives. 
In light of these results, we see our main contributions as:
\begin{itemize}
     \item Conducting the most robust application-grounded evaluation study of explainable ML methods to date that includes domain expert users, the intended task, real-world data and an inference strategy grounded in the operational goals. 

    \item Highlighting the critical importance of using evaluation metrics that capture operational goals, going beyond confusion matrix-based metrics and self-reported assessments of explanation quality.
    
    
    \item Illustrating the importance of designing experiments that reflect the deployment context and providing wide-reaching lessons for future evaluations of explainable ML methods in contexts beyond the setting of fraud detection by highlighting the contrasting conclusions between the earlier work and ours. 
    
    
    \item Highlighting the inefficacy of popular general-purpose explainable ML methods in our real-world setting, suggesting a need for developing methods targeted at specific use-cases
\end{itemize}



\section{Evaluation of Explainable ML methods}
\label{sec: background}
Evaluations of explainable ML fall into three categories: (1) functionally-grounded evaluation, where intrinsic qualities (e.g., fidelity to the underlying model) of the explanation are evaluated through algorithmic means, (2) human-grounded evaluation, where methods are evaluated with user studies but with simplified tasks, and (3) application-grounded evaluation where a method is evaluated with user studies involving domain experts in the intended deployment context \cite{doshi2017towards}. Application-grounded evaluations, though they enable assessing method utility in a real-world task, entail logistical challenges and complexities in executing trials directly on the use case \cite{chen2021interpretable,doshi2017towards,amarasinghe2021explainable}, and have remained elusive.   
Existing evaluation studies rely on proxies and simplifying assumptions such as focusing on the fidelity of explanations to the underlying model, relying on highly simplified versions of real-world datasets, and using readily available proxy users who lack domain expertise (e.g., workers on Amazon Mechanical Turk or users in academic settings)~\cite{islam2020towards,yalcin2021evaluating, ribeiro2016lime, ribeiro2018anchors, lundberg2018treeshap, lundberg2017unified, lakkaraju2016interpretable}. 
These simplified experiments fail to capture the complexities of real-world contexts; their results may overestimate the capabilities of explainable ML methods and compel practitioners to make implementation and deployment decisions based on  unreliable evidence of their effectiveness. 
Kaur et al. make a similar argument and claim that work in explainable ML has mainly focused on improving the explanation --- an artifact --- and is bound to fall short of improving outcomes based on human-ML collaboration \cite{kaur2022sensible}.
We summarize three main shortcomings of existing evaluation efforts that prevent practitioners from assessing a method's real-world utility:

\textbf{Performance on proxy tasks is not predictive of performance on real tasks.} 
The most common type of user study in the current explainable ML literature takes a human-grounded approach to evaluation, where the human performance on a proxy task is used as a metric of explanation utility \cite{doshi2017towards}. 
A commonly used example of such a proxy task ``forward simulation'', where a human is tasked with predicting the model's output given the input and explanation (e.g., as studied in \cite{hase_evaluating_2020, bell2022accuracy, lage2019human, lakkaraju2016interpretable, selvaraju2017grad, weitz2019do, lai2019on}). 
Although this task seems to capture a desirable property of helping users understand the inner workings of an otherwise black-box model, predicting the model output from input examples is rarely a task that individuals will face in a real-world setting. 
Therefore, it seems unlikely that this ability is either necessary or sufficient for explanations to assist end-users in decision-making. Bu\c{c}ina and colleagues \cite{buccina2020proxy} quantified this limitation, empirically showing that performance on a forward simulation proxy task aided by ML explanations was not predictive of the ability of those explanations to help users perform a real task. 

\textbf{User-reported, subjective measures of explanation quality are not reflective of explanation utility for the task at hand.}
Subjective measures such as user experience, preference, and trust are commonly used evaluators of explainable ML methods \cite{ribeiro2016lime, ribeiro2018anchors, selvaraju2017grad}.
Multiple studies have demonstrated that asking users what they \textit{think of explanations} is not indicative of the users' objective task performance when using those explanations. In \cite{green2019disparate, poursabzi-sangdeh2021manipulating}, authors showed that users' confidence in their decisions does not correlate with their performance on the outcomes of interest. Lakkaraju and Bastani found that explanations can manipulate users into trusting biased models \cite{lakkaraju2020how}. Bu\c{c}ina and colleagues \cite{buccina2020proxy} showed that subjective self-reported measures of explanation quality are not predictive of the performance on the task at hand.  

\textbf{Design flaws in the handful of existing evaluation studies in real-world contexts limit the conclusions that can be drawn.}
Application-grounded evaluations are rare in the existing literature. Even the handful that exists suffers from shortcomings in their experimental design.  
Lundberg and colleagues \cite{lundberg2018explainable} conducted a study to evaluate the utility of an ML system with SHAP \cite{lundberg2017unified} explanations to inform anesthesiologists during surgery with an experiment consisting of five anesthesiologists tasked with detecting hypoxemia risk using historical data collected from surgeries. 
The study showed compelling evidence that the ML system could improve decision-making in this high-stakes context. 
However, the experiment only compared the task performance of the physicians' aided by the explainable ML system to the physicians' performance based on their domain knowledge, failing to isolate the incremental effects of the explanations from the ML prediction/score by comparing the explainable ML system to the ML model alone.  
Therefore, it, unfortunately, provides no evidence for the \textit{incremental} utility of explanations themselves.
Jesus and colleagues \cite{jesus2021} experimented in the context of fraud detection, evaluating whether post-hoc explanations helped fraud analysts make better fraud detection decisions. The experiment consisted of three fraud analysts tasked with detecting fraud in e-commerce transactions. While they did run the experiment arms necessary to isolate the incremental effects of explanations, they made two major simplifying assumptions that diverged their experiment from the deployment setting. First, they resampled the data to a 50/50 distribution between fraudulent and non-fraudulent transactions, which deviated from the deployment setting. Second, they used metrics that did not align with the operational business goal to measure task performance, i.e., they used decision accuracy as the performance metric without taking into account the transaction value and the different costs of false negatives and false positives. While the authors attributed a statistically significant accuracy increase to the explainable ML methods, the simplifying assumptions make it unclear how well their results would generalize to a deployed system.


In response to the above shortcomings, several works have highlighted the need for more rigorous approaches for evaluating explainable ML methods in real-world contexts \cite{doshi2017towards, chen2021interpretable, amarasinghe2021explainable}. 
In particular, Amarasinghe and colleagues \cite{amarasinghe2021explainable} argue that four key elements are needed to adequately verify the real-world utility of a given explainable ML method: (1) \textit{A real task} that will be performed in the setting in which the explainable ML system will be deployed, including appropriate metrics to assess task performance that represent operational goals of the setting, (2) \textit{real data} collected from the deployment setting that reflects the nuances and complexities of the use case being considered, as model explanations deployed in practice will need to be robust to messy real-world data with complex relationships spanning thousands of predictors/features, (3) \textit{real users} who perform the task in the real world and have the domain expertise to not only make sense of but also use the model's explanations, additionally providing a realistic baseline that the ML-informed/augmented system needs to outperform, and (4) \textit{a robust inference strategy} to effectively evaluate the incremental impact of explainable ML methods, including sufficient sample sizes for statistical power, appropriate hypotheses and experimental conditions, and valid analytical methodology to capture uncertainty in the data and support conclusions being drawn.



\begin{table*}[t!]
  \caption{A summary of our modifications to the setup of Jesus et al. \cite{jesus2021} to bring the experimental setup closer to the deployment context.  
  }
  \label{tab:study-diffs}
  \def\arraystretch{1.2} 
  \begin{tabularx}{\linewidth}{ l X X }
    \hline
     & Jesus et al. \cite{jesus2021} & This Study \\ 
    \hline
     Label Positive Base Rate & 50\% (resampled to artificially balance the data ) & 15\% (actual rate) \\
     Evaluation Metric & Accuracy (and related confusion matrix-based metrics) & Percent Dollar Regret (PDR) reflecting the domain-specific goal/metric\\
     Interface & Accept or Reject Transactions & Accept, Reject, or Escalate Transactions\\
     Experimental Arms & Data, Model, SHAP, TreeInterpreter, LIME & Data, Model, SHAP, TreeInterpreter, LIME, Random Explanations, Irrelevant Explanations\\
     Sample Sizes (n) & 200 (controls), 300 (explainers) & 500 (all conditions) \\
     Post-Decision Questions & Specific to explainers & Comparable across all arms\\
     Follow-Up Interviews & Not Performed & Conducted with all analysts\\
  \hline
\end{tabularx}
\end{table*}

\section{Designing an application-grounded evaluation study}

We focused on designing an experiment that replicates the objectives of the real-world context in which explainable ML systems are deployed, thereby relaxing the simplifying assumptions typically made by studies evaluating explainable ML methods. 
In particular, our work expands and improves upon a previous study \cite{jesus2021} that studied the utility of ML explanations in a fraud detection setting. 
In this section, we first introduce the application context, and then, elaborate on the improvements we made to the experimental setup (summarized in Table~\ref{tab:study-diffs}) to relax the simplifying assumptions and make the setting consistent with the deployment context. 

\subsection{Application Context: Credit Card Fraud Detection}

The fraud detection task in this work is common among online retailers, financial services, as well in healthcare contexts. 
In such a setting, each transaction is scored by an ML model for its risk of being fraudulent. 
Transactions with very low risk are automatically approved, those with very high risk are automatically declined, but those with intermediate risk scores are flagged for review by human analysts. 
We partnered with the same large e-commerce merchant as the authors of~\cite{jesus2021} (company name withheld in order to discuss the details of the results), making use of the fraud risk model that is currently deployed in their production system. 
For this client, fraud risk is calculated using a Random Forest model. Scores are scaled to range from 0-1000, with auto-approval of transactions with a score under 500 (82.1\% of all transactions) and an auto-rejection threshold of 617 (3.2\% of all transactions). 
We study the set of transactions with scores between 500-617, which fall into the \textit{review band} and are reviewed by fraud analysts.
It is worth noting that a small proportion of transactions outside of this score range are also flagged for review based on client-specific heuristics, representing 6.2\% of transactions in our dataset.
The set contains data from 231,362 historical transactions that took place between 26 Sep 2019 and 20 Nov 2019. We recruited the same fraud analysts who participated in the study of \cite{jesus2021}. However, it is worth noting that the analysts were not shown any transaction they had seen before. 

\begin{figure*}[t]
   \subfigure[]{\includegraphics[width=0.50\linewidth]{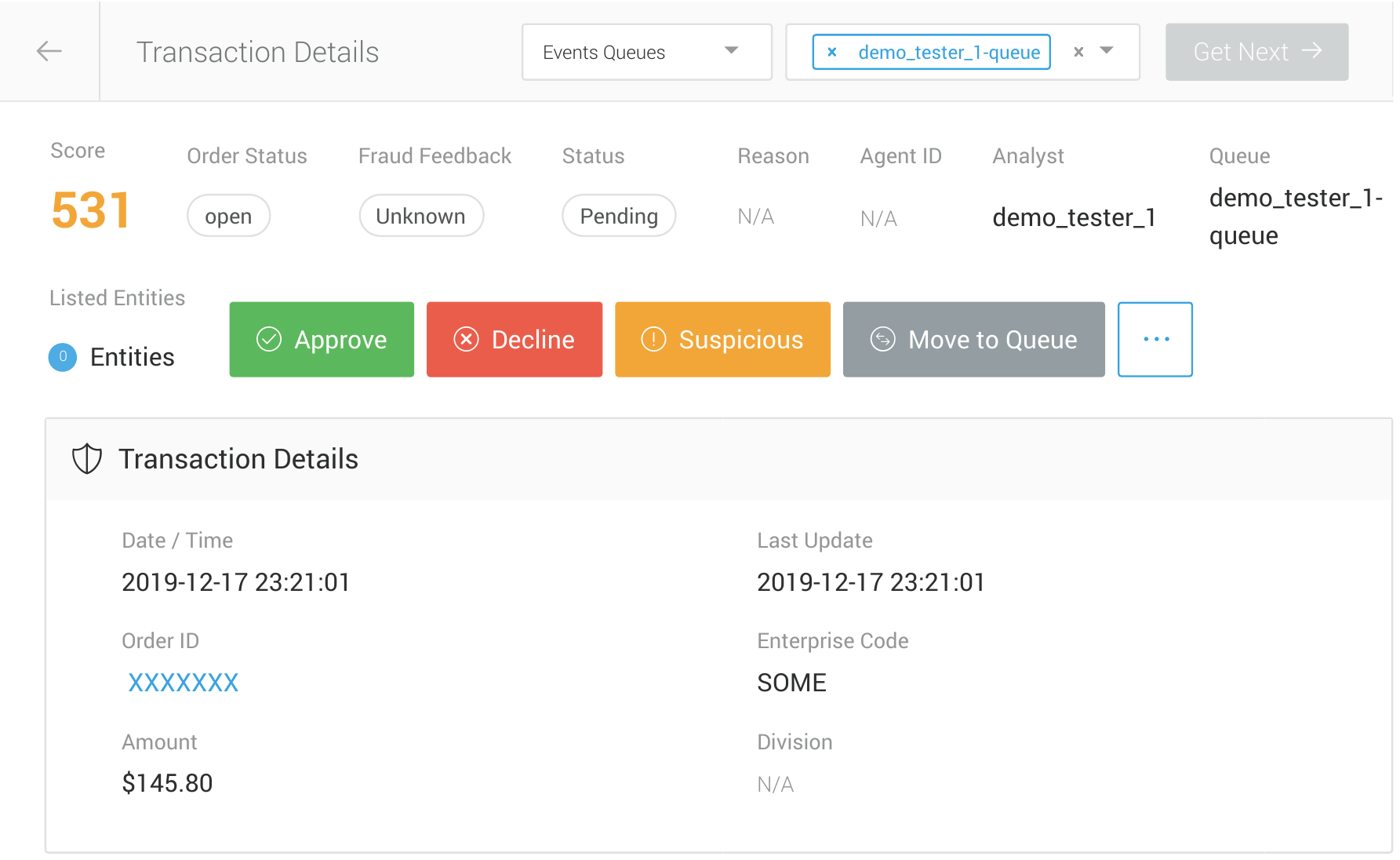}}
   \subfigure[]{\includegraphics[width=0.49\linewidth]{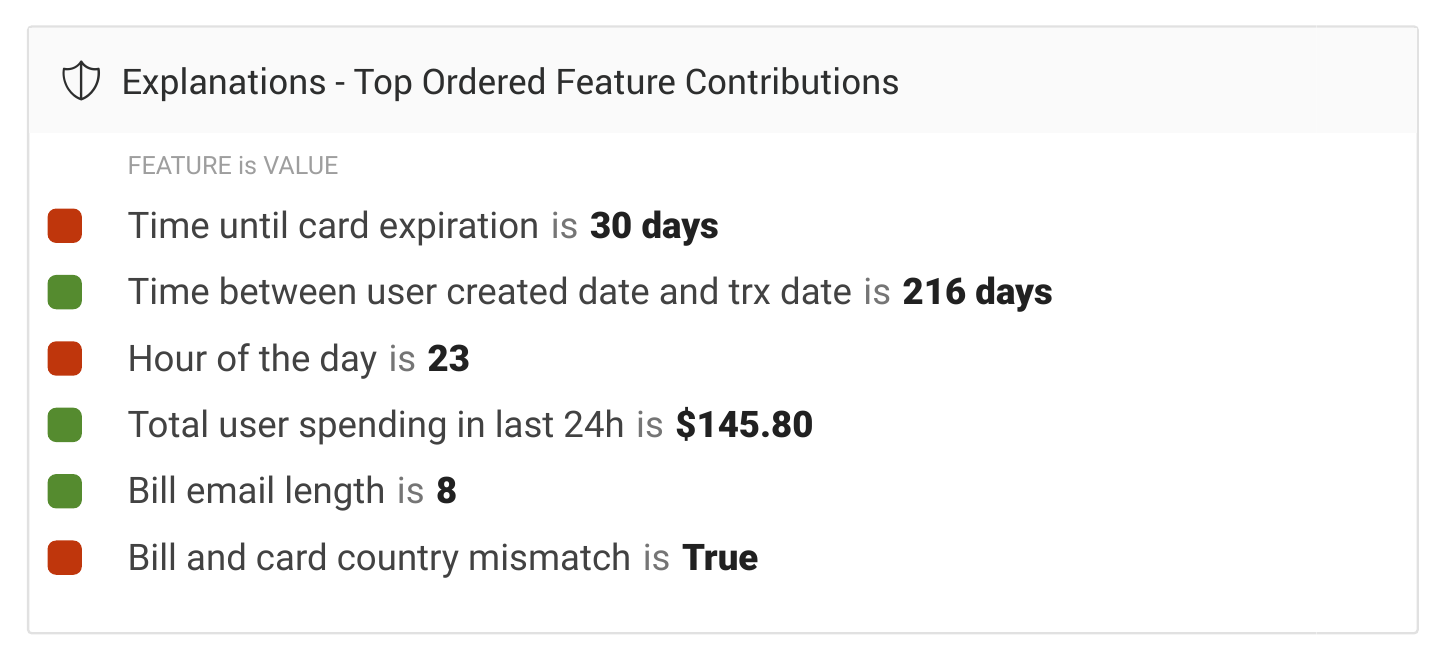}
   \label{fig:screenshot_explanation}}
    \caption{Example of the user interface used by the analysts in the experiment (populated with sample data for illustrative purposes). (a) Basic interface components, including the model score, labeling buttons, and transaction details (see Appendix Figure \ref{fig:full_interface} a sample of the full interface and all components). (b) A component of the interface containing explanations of the model score which was included under transaction details when present.}
    \label{fig:interface_screenshot}
\end{figure*}

 Transactions in the review band are presented to an analyst with detailed information including billing and shipping addresses, items being purchased, client-side information (e.g., geolocation of the IP address, type of browser and device, etc), and the history of the credit card or user account associated with the purchase. Figure~\ref{fig:interface_screenshot}(a) shows an example of the interface the analysts use for this task. Based on the provided information, the analyst must choose to either: 
\textit{allow} the transaction to proceed, generating revenue if the transaction is legitimate but risking approving fraudulent ones;
\textit{reject} the transaction to avoid fraud, but risking blocking a legitimate transaction;
or, \textit{escalate} the transaction for further review by a more senior analyst signaling uncertainty. 
This additional review typically entails further research or contacting the customer directly to confirm that they are attempting to make the purchase. 
While this escalation is much more likely to arrive at the correct decision of the transaction, it is also far more costly in terms of the time and resources involved to reach the decision.

\subsection{Making the experiment setup consistent with the deployment context}
\label{sec: modifications}
Our work improves upon a previously conducted explainable ML evaluation study \cite{jesus2021} on the same fraud detection setting. These improvements (summarized in Table~\ref{tab:study-diffs}) include 1) modifying the evaluation metrics to reflect the operational business objective, 2) correcting data distribution errors, and 3) other improvements to the experimental design, yielding an experiment setup that reflects the actual deployment setting more closely and is better suited to evaluate the utility of the post-hoc explanation methods considered. We now discuss each improvement in more detail:

\subsubsection{Performance metrics that reflect operational goals:}
In reality, fraud analysts' decisions are measured on two dimensions: (1) decision correctness (and related revenue impacts) and (2) decision speed. 
One crucial simplification made by the authors of the previous study is in how they measured decision correctness: they assume that focusing on traditional confusion matrix-based metrics (i.e., accuracy, false positive rate (FPR), precision, and recall) would suffice to represent task performance and demonstrate whether explainable ML methods meet the fraud-detection needs of the business.
However, such metrics do not capture the operational goals of the business. In particular, confusion matrix-based metrics ignore the dollar value of the transaction (i.e., assumes all transactions are equally important to the merchant) and the relative importance of the two types of errors (i.e., assumes that both types of errors have the same impact on the merchant).

The concept of ``correctness'' depends on the relative costs of false negative and false positive errors to the business objectives (i.e., a more nuanced quantity than overall ``accuracy''). 
Here, we formalize correctness through a utility metric we refer to as \emph{Percent Dollar Regret} (PDR), which captures the revenue lost due to incorrect decisions relative to what would be realized if all the reviewed transactions were perfectly classified. 
The PDR metric reflects an accounting of the relative cost/benefit of each decision based on the value of the transaction and where it falls in the confusion matrix. 
To capture these nuances, we weigh each decision/transaction by a coefficient that captures both above factors (with these coefficients applied to the dollar value (in USD) of the transaction, $v_i$ ). The applied coefficient is based on where a decision falls in the confusion matrix and its downstream impacts on the business goals: 

\begin{itemize}
    \item \textit{True Negative (TN)}: A legitimate transaction is correctly identified and let through. The merchant would gain the revenue from the sale, and any future worth the customer creates for the merchant. The merchant would gain the revenue from the sale, and any future worth the customer creates for the merchant. Therefore, we weigh TNs with the coefficient $ ( 1  + \lambda ) $, where $ \lambda $ is the expected value for the long-term worth the customer would generate as a multiple of the current transaction value. 
    \item \textit{True Positive (TP)}: A fraudulent transaction is correctly identified and blocked. The merchant will neither lose nor gain any revenue. Hence, the weight of a TP is $ 0 $.
    \item \textit{False Negative (FN)}: A fraudulent transaction is incorrectly classified as legitimate and let through. The real owner of the credit card would dispute the charge and the money would be returned. The merchant would lose the item and  have to pay a surcharge to the credit card processors. 
    Additionally, allowing fraudulent transactions incurs two longer-term risks: first, perpetrators of fraud can discover the lax practices and draw more fraud to the site, and second, high fraud rates can lead to penalties from credit card processors. 
    Together, these long-term effects increase the relative costs of false negatives as well as introduce a level of risk aversion to allowing fraud to go undetected.
    We capture these combined short- and long-term costs with a weight parameter, $ \alpha $
    \item \textit{False Positive (FP)}: A legitimate transaction is marked as fraud and is blocked. The merchant runs the risk of losing the sale, and losing the customer and their future, long-term revenue. We weigh a FP with the weight $ (1 - \beta) + (1 - \delta) * \lambda $. Both losses are subject to a probability: in the short-term, customer can use a different payment method to complete the purchase and $ \beta $ refers to the probability of losing the current sale; and in the longer term, the negative experience of having transactions rejected may lead customers to move to other online merchants, which we capture with probability $ \delta $ of losing the long-term expected value $ \lambda $. Therefore, we consider that with FPs, some revenue is generated albeit with a lower expected value.
\end{itemize}

Based on these, we calculate the PDR metric over a set of transactions (and the corresponding customers) in the following form: 

\begin{equation}
  PDR =  1-\frac{\textrm{Realized~Revenue}}{\textrm{Possible~Revenue}} = \\ \frac{\sum_i(
  \mathbb{1}(y_i=0, \hat{y}_i=1)*(\beta+\delta*\lambda)
+ \mathbb{1}(y_i=1, \hat{y}_i=0)*\alpha)*v_i}{\sum_i\mathbb{1}(y_i=0)*(1+\lambda)*v_i}
\end{equation}

where $\mathbb{1(\cdot)}$ is an indicator function that takes value 1 if the argument is satisfied and 0 otherwise, $y_i$ is the actual label of transaction $i$ (1 indicating fraud and 0 indicating legitimate), $\hat{y}_i$ is the label assigned by the analyst for transaction $i$, $v_i$ is the value (in US Dollars) of transaction $i$, and each sum is taken over all transactions (the other parameters are as described above). 
The denominator here is the expected revenue from all legitimate transactions (regardless of how they are marked by the analysts) while the numerator sums up the expected revenue \textit{lost} when the analysts are incorrect in either direction. 

An additional factor to consider is the impact of \textit{escalation} decisions on decision time and correctness. When an analyst marks a ``suspicious'' transaction, it is elevated to a senior analyst who takes a deeper dive into the transaction, incurring a time cost. Therefore, for each decision escalated, we assume that the senior analyst arrives at the correct decision but we apply a time penalty ($ \tau $).

\subsubsection{Ability to escalate ``suspicious'' transactions}
Analysts in the previous study could only decline or approve each transaction, but in the \textit{real world}, when evaluating live transactions, they have a third option, which is to escalate suspicious transactions for further review. To better reflect the true deployment context --- and because this escalation rate may be an interesting outcome in its own right --- the present study provided analysts with all three options.

\subsubsection{Evaluating additional hypotheses with additional experiment arms}
Although we focus on the same set of post-hoc explanation methods here as in the previous work to support comparison, we added two additional control/placebo arms --- one where we show random explanations to the analysts, and another where we show completely irrelevant explanations to better evaluate whether any observed impact of explanations was due to the quality of the explanations themselves or simply the mere presence of an explanation.  

\subsubsection{Post-decision Questionnaire}
 In the previous work, analysts were asked a set of post-decision questions about the relevance and usefulness of ML explanations only after explanation arms. To provide a better baseline, our study asked a uniform set of questions in every experiment arm: the analysts were asked to indicate their confidence in their decision, the quality of the available information, and the data elements they used to inform their choice.

 \subsubsection{Conducting Post-experiment Analyst Interviews}
 Following the completion of data collection for the experiment, we conducted qualitative interviews with each of the analysts to better understand how they perceived the explanations, the balance they tried to strike between speed and correctness, and the extent to which the experimental setting may have departed from the day-to-day evaluation of live transactions.

 \subsubsection{Making the sample sizes larger for increased statistical power}
We increased the power of the analysis by using 500 transactions per experimental arm (compared to 200 in control arms and 300 in explainer arms in the previous study).

\subsection{Experimental Hypotheses and Arms}
To understand the incremental impact of additional information (including raw transaction details, ML model risk scores, and explanations of those scores) on analyst decisions, we designed our experiments to evaluate five hypotheses: 

\begin{enumerate}
    \item[H1] \textit{Having access to the model score improves analysts' decision-making performance compared to only having access to the transaction data}. While our primary goal is to evaluate the incremental impact of post-hoc explanation methods, we argue an important first step is to evaluate the impact of just the ML model predictions on analyst performance.
    \item[H2] \textit{Having access to an explanation of the model score improves analysts' decision-making performance compared to only having access to the model score and data}. 
    Here, we are interested in identifying any incremental impact of showing analysts the explanations of the predictions. 
    We hypothesize that the explanations can help the analysts focus their attention on relevant pieces of data. help them confirm or override the model predictions, and improve their performance. 
    \item[H3] \textit{Explanations make the analysts more confident in their decisions, resulting in fewer transactions being escalated to a senior analyst}. 
    We are additionally interested in learning whether the analysts become more confident/decisive when they are presented with model explanations. If this is true, we should observe a reduction in the number of transactions escalated  (marked as \textit{suspicious}).
    \item[H4] \textit{The impact of explanations on decisions is different based on which post-hoc explainer is used}. Different post-hoc explanation methods in general do not produce the same explanation 
    for the same prediction and same model, and thus we hypothesize that explanations generated from different methods would have different impacts on the analysts' decisions. 
    \item[H5] \textit{Explanations generated from an ad-hoc method would be worse compared to those generated by ``bona fide'' explanation methods}. 
    We evaluate whether any effects we see with the explanations are attributed to the mere presence of an explanation (which could affect how the analysts interact with the task, perhaps prompting them to be more thoughtful about their decision regardless of the quality of the explanations themselves), or due to the explanation content and the performance of the method that generated it.   
\end{enumerate}

We study the same post-hoc explanation methods as \cite{jesus2021} --- LIME \cite{ribeiro2016lime}, TreeSHAP \cite{lundberg2018treeshap}, and TreeInterpreter \cite{saabas2015}.
All methods are feature-attribution, i.e. explanations consist of feature name and feature-importance score pairs. 
Similar to the previous study, we show the top 6 features (i.e., the 6 features with the largest absolute importance) to the analysts.  
As the feature importance values can be positive or negative, the polarity of the feature's importance is represented using green / red colors as shown in Figure~\ref{fig:screenshot_explanation}. 
Green indicates that the feature importance is negative (pulling the score away from fraud), and red indicates positive importance values (pulling the score toward fraud).
To evaluate the hypotheses for the explanation methods, we organized our experiment into seven different experiment arms:

\begin{itemize}
    \item \textit{Data}: The fraud analysts only see the raw data pertaining to the transaction and any
history that is available. Serves as the control for H1. 
    \item \textit{ML Model}: The fraud analysts see the fraud score predicted by the ML model in addition to the raw transaction data. Serves as the treatment arm from H1, and the control for H2 and H3. 
    \item \textit{TreeSHAP}: In addition to the transaction data and the ML model fraud score, the analysts see a post-hoc explanation generated using TreeSHAP \cite{lundberg2018treeshap}. Serves as the treatment arm for H2, H3, H4 and also as the control for H4 as we compare across explanation methods 
    \item \textit{TreeInterpreter}: Similar to \textit{TreeSHAP}, but the post-hoc explanations are generated using the method TreeInterpreter \cite{saabas2015}. Serves as the treatment arm for H2, H3, H4 and also as the control for H4 as we compare across explanation methods  
    \item \textit{LIME}: Analysts see post-hoc explanations are generated using the method LIME \cite{ribeiro2016lime}, in addition to transaction data and the ML model score. Serves as the treatment arm for H2, H3, H4 and also as the control for H4 as we compare across explanation methods  
    \item \textit{Random}: The analysts see an explanation in addition to the data and ML model, but the explanations contain random features from the larger feature set, as opposed to explanations from a ``bona fide'' explainable ML method. Serves as the control for H5 
    \item \textit{Irrelevant}: Analysts see explanations consisting of ``irrelevant'' features (e.g., seconds component of the transaction timestamp, last two digits of the IP address, etc.). See the Appendix Table \ref{tab:irrelevant_features} for a complete list of irrelevant features. Serves as the control for H5. 
\end{itemize}

Three professional fraud analysts familiar with the reviewing system participated in the experiments. Since we were limited to three analysts, we were unable to randomize across users to create our treatment and control/placebo arms. 
Instead, we randomized across transactions\footnote{It is worth noting that while we used the same set of transactions as the previous study, we randomized the samples across users so that the users were seeing them for the first time.
} and organized the experiments in three stages. In Stage one, we only showed the transaction data, in Stage two we introduced the ML-based fraud score, and in Stage three, we introduced explanations and conducted all the explanation experiment arms in sequence.

\section{Key Findings}

\begin{table*}[t]
  \caption{Performance summary across the experiment arms}
  \label{tab:summarytab}
  \begin{tabularx}{\linewidth}{l X X X X X X X X X}
    \hline
    Arm & PDR & Time & Acc & FPR & TPR & Prec & Appr. & Decl. & Escl. \\
    \hline 
     Data Only & 9.5 & 79.2 & 76.6 & 18.7 & 48.6 & 30.7 & 71.7 & 22.4 & 5.9\\
     Model &  8.9 &  51.3 &  82.2 &  12 &  49.3 & 42 &  80.8 &  17.0 &  2.2\\
     TreeSHAP  &  9.7 &  67.3 &  81.9 &  10.6 &  38.4 &  38.4 &  83.1 &  13.7 &  3.2 \\
     TreeInterpreter  &  10 &  65.6 &  80.9 &  12.1 &  40.5 &  37 &  81.7 &  15.5 &  2.8 \\
     LIME      &  11.6 &  57.7 &  83.2 &  8 &  38.7 &  43.3 &  85.2 &  12.2 &  2.6\\
     Random Exp.    &  9 &  56.3 &  82.7 &  10 &  38.7 &  42 &  85.5 &  13.3 &  1.2 \\
     Irrelevant Exp. &  9.7 &  62.2 &  81.2 &  9.5 &  29.9 &  36.5 &  85.8 &  12.6 &  1.6 \\
  \hline
\end{tabularx}
\end{table*}

\begin{figure*}[t]
  \centering
  \subfigure[]{\includegraphics[width=0.48\linewidth]
  {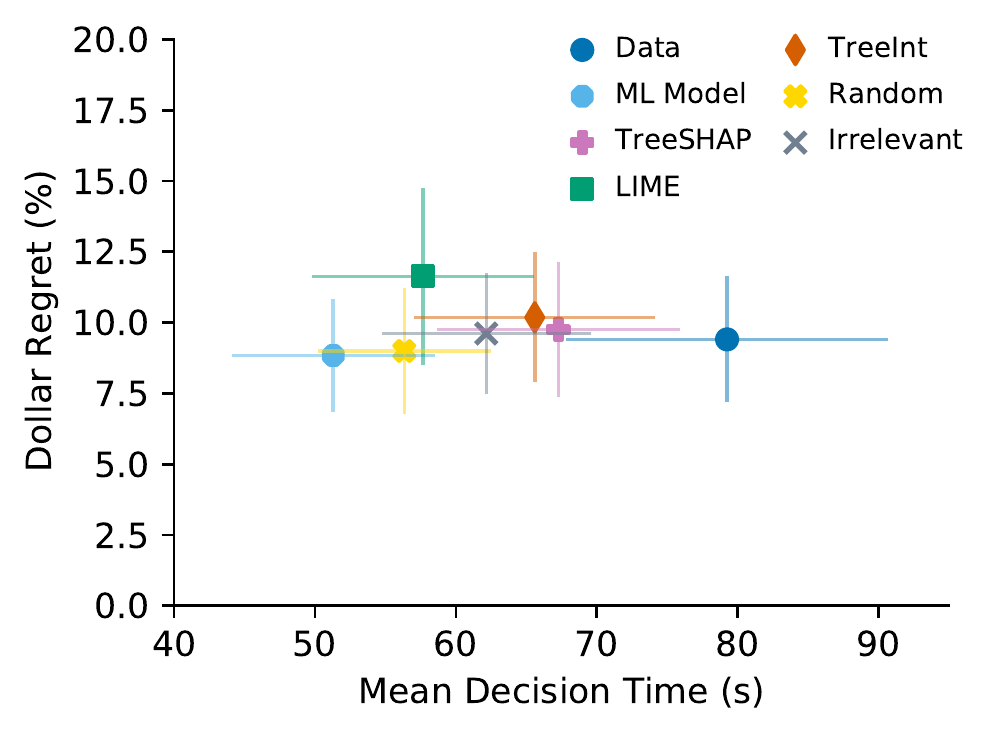}
    \label{fig:pdrtime_a}}
  \hfill
  \subfigure[]{\includegraphics[width=0.48\linewidth]{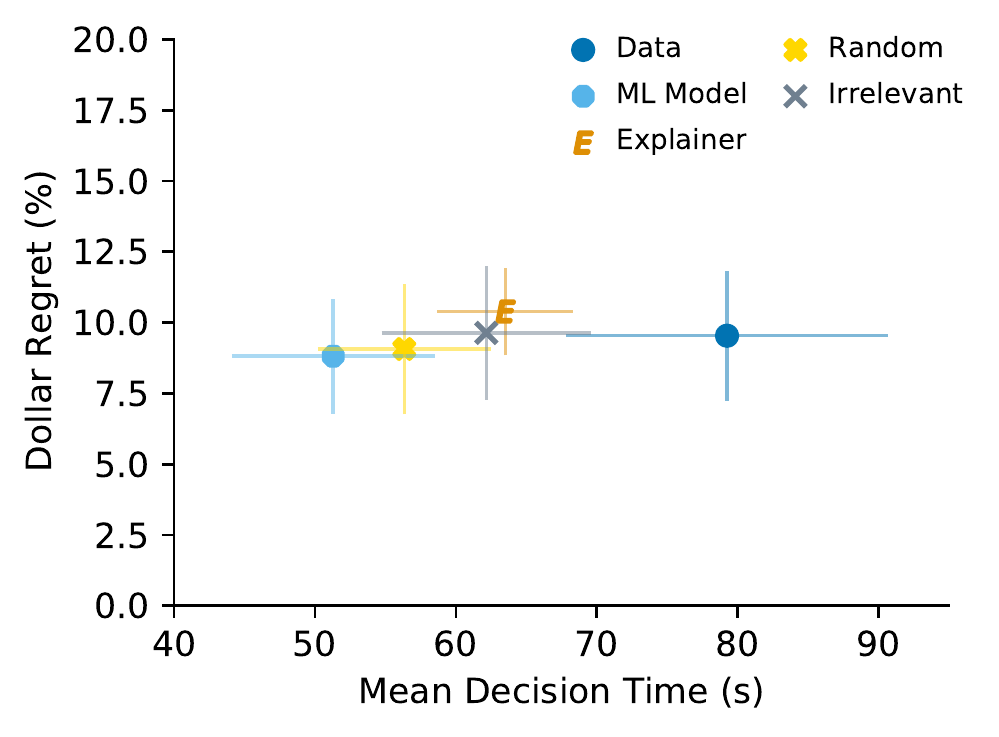}
   \label{fig:pdrtime_b}}
    
    \caption{
    PDR metric versus Mean Decision Time in the experiment arms.  While the PDR metric remains similar across arms, we observe that introducing the ML score makes the analysts faster, whereas introducing explanations slows them down without improving performance. The error bars show the 90\% CIs. (a) The individual experiment arms. (b) Explanation methods grouped together.
    }
    \label{fig:pdrtime}
\end{figure*}
 \vspace{-0.2cm}

\begin{figure*}[h]
   \centering
   \subfigure[]{\includegraphics[width=0.48\linewidth]{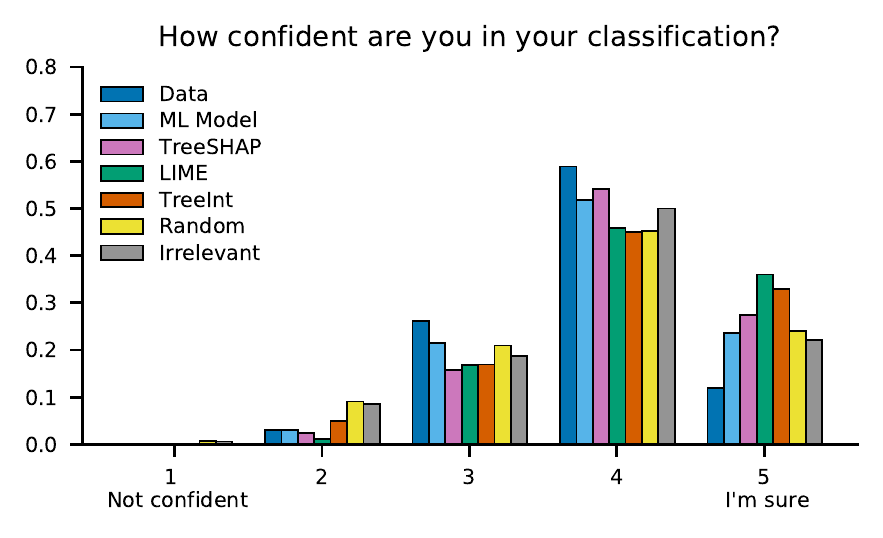}}
  \hfill
   \subfigure[]{\includegraphics[width=0.48\linewidth]{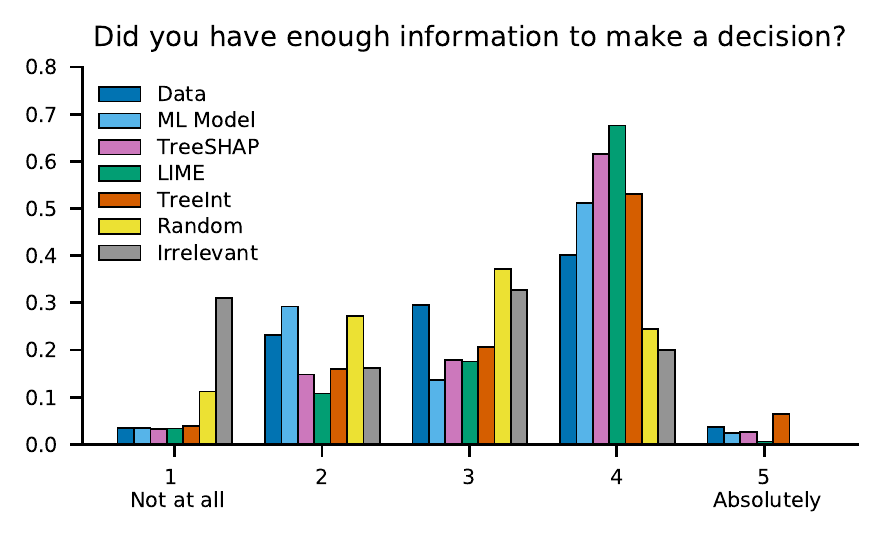}}
    \caption{
    Analyst responses to post-decision questions for rating their confidence and perceived quality of available information. Ratings for confidence and information quality increase as the experiment progresses through the three stages, but without any correlation to their behavior or performance. Users respond with lower ratings for both questions for Random and Irrelevant arms, however, without a change in behavior or performance. (a) Analysts' confidence in their decision. (b) Perceived quality of the available information.}
    \label{fig:questionnaire_results_sus}
\end{figure*}

We analyzed the performance of the analysts in each experiment arm using two metrics, PDR and the mean decision time. We initialized the parameters for metric calculation with $ \alpha = -3$, $\beta = 0.5$, $\delta = 0.1$, $\lambda = 3 $ and $ \tau = 600s $ based on data analysis and consulting the analysts post experiment completion. In evaluating our experiment hypotheses, we employed group mean comparison with statistical tests and generalized linear models (GLMs) to account for the differences in analyst performance. We found that our findings were consistent across the two analyses, and we provide more details about hypothesis tests and parameter initialization in the Appendix Extended Methods section.
Figure \ref{fig:pdrtime_a} shows the PDR metric against the average decision time for all the experiment arms, while Figure \ref{fig:pdrtime_b} shows the same with all explainers combined into one group. We also report decision accuracy metrics for completeness and additional context, as shown in Table \ref{tab:summarytab}.

\subsection{Finding 1: The ML model helped the analysts make faster decisions}
\label{sec: impact_model}
An important baseline for understanding the impact of ML explanation methods is understanding how the ML model scores by themselves affect analyst decisions, which we explored via hypothesis H1.
To evaluate H1, we compared performance between \textit{Data} and \textit{ML Model} arms.
Analysts became significantly faster with the \textit{ML Model} compared to \textit{Data}, reducing the average decision time by almost 30 seconds $ (p=.0001)$ (see Figure \ref{fig:pdrtime}).
Although we did not observe significant improvements in PDR with the \textit{ML Model}, we found appreciable improvements on several confusion matrix metrics: decision accuracy improved significantly $ (p=.03) $, FPR decreased by 6 percentage points, and Precision increased by 9 points (See Table \ref{tab:summarytab}). 
The significant increase in accuracy is a result of analysts rejecting significantly fewer transactions $ (p=.03) $ without decreasing TPR and approving significantly more transactions $ (p=.0008) $ while reducing the FPR.
While these improvements in confusion matrix metrics may seem at odds with the observed lack of change in the PDR, the difference can be explained by the fact that PDR captures the relative costs of FN and FP errors. 
As we mentioned, the merchant is more risk-averse towards FNs than FPs, meaning that the PDR metric will be more sensitive to reductions in FNs rather than FPs. Since the \textit{Data} and \textit{ML Model} arms have similar FN rates, the improvements in accuracy almost entirely resulted from trading FPs for TNs, thereby muting the extent to which the accuracy improvement translates into a corresponding PDR improvement (with the 6pp increase in accuracy yielding only a statistically insignificant 0.6pp drop in PDR). 
The contrast between these two results reinforces the importance of choosing evaluation metrics that appropriately reflect operational objectives. In particular, assuming that different types of errors have the same cost can yield misleading results.

We also find the \textit{ML model} made the analysts more confident, resulting in a significant decrease in the escalated transactions cutting the escalation rate by more than half $(p = .003)$.
It is worth noting that the reduction in the escalation rate is one significant factor contributing to the improvement in decision speed.  
The analysts also showed significantly higher levels of self-reported confidence with the \textit{ML model} compared to \textit{Data} (with the analysts being sure of their decisions 24\% of the time compared to 12\%, respectively) $ (p < 10^{-5}) $. 
Similarly, they reported a significant increase in the perceived quality of information with the \textit{ML model} , with analysts reporting they had high quality information (ratings of 4 or 5 out of 5) 54\% of the time compared to 43\% of the time with \textit{Data} alone $(p=.002)$.  

\subsection{Finding 2: Post-hoc explainers did not show significant incremental impact}
To evaluate H2, we compared each \textit{Explanation} arm and a \textit{Combined Explainer}---an arm with all the explanation methods grouped together---to the \textit{ML model} arm.
We found no significant differences in PDR, or any confusion-matrix-based metrics between any individual \textit{Explanation} arm and \textit{ML Model}.
However, the explanations did make the analysts slower, with statistically significant differences between the \textit{ML Model} and \textit{TreeInterpreter} $ (p=.03) $; \textit{ML Model} and \textit{TreeSHAP} $ (p=.01) $; and \textit{ML Model} and the \textit{Combined Explainer} $ (p=.03) $.
As the speed of decisions directly impacts business utility, the explanation methods worsened the outcomes of interest.
With respect to H3, we did not observe any statistically significant differences in the escalation rates between the \textit{ML Model} and any \textit{Explanation} arm. 
We did observe \textit{Explanation} arms reduced the decline rate  compared to the \textit{ML Model} $ (p=0.03) $---with the exception of \textit{TreeInterpreter}---without impacting any outcomes of interest. 
Further, we compared \textit{Explanation} arms against each other to evaluate H4 and, again, did not observe any statistically significant differences in any metric, suggesting that all the tested explanations negatively impact analyst decision speed without improving decision-making. 
Nevertheless, the analysts reported \emph{higher} confidence in their decisions with the \textit{Explanation} arms compared to the \textit{ML Model} (Figure \ref{fig:questionnaire_results_sus}). 
Analysts reported they were sure of their decision (rating of 5) 36\% of the time for \textit{LIME}, 33\% of the time with \textit{TreeInterpreter}, and 27\% of the time with \textit{SHAP}, compared to the 24\% with the only the ML model. 
\textit{LIME} $ (p < 10^{-4}) $  and \textit{TreeInterpreter} $ (p < .001) $  showed statistically significant differences. 
Likewise, analysts reported a higher perceived quality in information with all explanation methods, compared to the \textit{ML model}. The analysts assessed the provided information as high quality (rating of 4 or 5) 68\% of the time with \textit{LIME} $ (p < 10^{-5}) $, 64\% of the time with \textit{TreeSHAP} $ (p < 10^{-3}) $, and 60\% of the time with \textit{TreeInterpreter} $ (p=.06) $, compared to the 54\% with the \textit{ML Model}. 
It is worth noting that this comparative analysis was not possible in Jesus et al. \cite{jesus2021}, which only included a questionnaire to study the quality of explanations.

\subsection{Finding 3: Explanations with irrelevant/random features has similar performance to post-hoc methods}
We ran the \textit{Random} and \textit{Irrelevant} experiment arms to evaluate H5 and found that  both resulted in very similar PDR  and confusion matrix metrics compared to the \textit{ML model} and \textit{Explanations}. 
We noticed that irrelevant explanations significantly slowed down the analysts compared to the ML model score $ (p=.08 ) $, but the decrease was similar to that incurred by \textit{real} explanation methods.
Interestingly, \textit{TreeSHAP} made the analysts slower than Random explanations $(p=.08)$ \emph{and} showed significantly higher escalation rate $(p=.03)$. 
We could attribute these differences to \textit{TreeSHAP} being the first \textit{Explaination} experiment arm, 
while \textit{Random/Irrelevant} was the last, so the analysts seemingly got better at ignoring explanations (which were not helping them make better decisions). 

Analysts reported significantly lower confidence on both \textit{Random} ($p < .002$ for all tests) and \textit{Irrelevant} explanations ($p \leq .05$ for all tests) compared to all the post-hoc explanation methods, with decision confidence ratings comparable to the \textit{ML Model} arm. 
(Figure \ref{fig:questionnaire_results_sus}). 
The analysts also reported drastically lower perceived quality of information, even when compared with the \textit{Data} and \textit{ML Model} arms $ (p=0)$, despite the fact that the analysts still had access to the same raw data and model score. 
Surprisingly, these decreases in self-reported confidence and information quality did not translate into a higher escalation rate (Table \ref{tab:summarytab}), suggesting that although the subjective, self-reported metrics could indicate the analysts' ability to discern between reasonable and unreasonable explanations, the analysts were unable to translate their beliefs into effects on the true operational metrics of interest.

\section{Discussion}

\subsection{ML models help but ``explanations'' do not provide incremental benefit for this task} 
Our findings suggest more utility for ML models than post-hoc explanations in aiding analyst decision making. The lack of improvement in the PDR metric across arms seems surprising but perhaps suggests that the models are picking up on the same signals in the data as the analysts themselves would. 
Thus, the scores do not provide novel information to help the analyst discriminate between legitimate and fraudulent transactions. 
The model does appear to, however, help the analysts more efficiently process transactions without reducing task performance. The analyst can spend less time drawing the \emph{same} conclusion by relying on the model. Similarly, providing the model score reduced the escalation rate by 63\%, further improving the decision time. 
In contrast, although we had hypothesized that explanations would make the analysts both faster and more accurate, the results here provide considerable evidence to the contrary. 
Rather than helping the analysts focus their attention on the most relevant data elements, it seems the explanations simply added additional elements on the interface for the analysts to spend time considering, without providing useful new information to help them improve on their task.

\subsection{Increases in self-reported confidence do not reflect changes in behavior or performance}
Despite their failure to improve actual performance metrics, the explanations did have an impact on the analysts' self-reported confidence in their decisions and their assessment of the quality and sufficiency of available information. 
However, this increase in perceived information quality and reported confidence did not result in a change of analyst behavior.
For instance, despite the increase in subjective confidence metric, the objective measure of confidence --- the escalation rate --- did not change.
This disconnect between user assessments of explanation utility and the lack of evidence that those explanations improved analysts' performance at the task highlights the importance of focusing on practical, application-grounded evaluations instead of asking users how much they like or trust the explanations.
Our finding aligns with similar observations by both Hase and Bansal \cite{hase_evaluating_2020} and Bu\c{c}ina et al \cite{buccina2020proxy} that there was little to no relationship between subjective and more practical metrics in their evaluations of several classes of explainers.

\subsection{Experimental design affects conclusions}
Our findings pose an interesting contrast with the prior study in the same fraud-detection setting \cite{jesus2021}. 
Findings of \cite{jesus2021} suggested analysts were faster but much less accurate with model scores compared to data alone, but much of this lost accuracy was recovered upon the introduction of post-hoc explanations at the expense of the loss of some of these speed improvements. 
Although our current results display the same pattern with respect to decision times (analysts are slowest with data alone, fastest with model scores, and intermediate with explanations), our findings on accuracy do not match the prior study: we see improvements in accuracy with the model but no difference between model and explanations.
The divergence in the findings between these two studies provides evidence that seemingly small experimental design choices (as summarized in Table~\ref{tab:study-diffs}) can have significant impacts on the results. Ensuring that the experimental context, data, tasks, and evaluation metrics reflect real-world settings in which these tools will be deployed is crucial to drawing conclusions from the results that will generalize to the systems in production.

We describe a few aspects of the experimental design that may have led to the differences in conclusions: First, because our study uses the observed base rate of positive labels in the data (15\%) rather than an (artificial) equal balance of fraudulent and legitimate transactions as in the previous study, using \textit{accuracy} as a statistical metric becomes inherently less meaningful. 
Second, the more realistic base rate in the data likely affects the prior beliefs the analysts operate under when participating in the experiment. For instance, if the analysts have a lower expectation for the fraud rate in the review band, asking them to evaluate transactions under a balanced distribution could contribute to degraded performance.
Beyond statistical accuracy itself, our present study also accounts for the considerably different costs of false positives and false negatives by focusing on the PDR metric to more appropriately reflect the operational goal, finding that improvements in accuracy do not necessarily translate into improvements in PDR. 

\subsection{Analyst interviews highlighted the mismatch between their needs and explainable ML methods}
After completing the experiment, we interviewed the analysts who participated in the study through informal discussions. 
We wanted to understand how they approached the experiment to verify that their approach matched the deployment process. Further, we wanted to understand how the analysts used explanations in their decisions and how we design new ML explanations to assist them in their tasks. 
Through these interviews, we learned that the analysts approached the transactions in the experiments as they would with a client, verifying the consistency of the experiment with the deployment context. Further, the analysts indicated that they prioritized both correctness and decision speed, verifying our performance metrics that captured both aspects.

While they mentioned that explanations helped them identify parts of the transaction data to focus their attention on, the analysts suggested that there was considerable room for improvement in the additional information beyond the model scores that would be most helpful in the decision-making. 
For instance, they believed that augmenting the feature attribution explanations with more context could help build trust over time, e.g., how the highlighted features align with their/organization's past decisions and whether they reflect accurate information. 
Further, the analysts suggested using the explanations to modify the interface could significantly impact performance by helping the analysts focus on the pertinent information, saving the time it would take to scan the entire transaction. 
This feedback further emphasizes the need for designing both explainable ML methods and human-computer interfaces that address specific requirements of use cases.

\section{Limitations and Future Work}
Although we sought to carefully design an experiment that as closely reflects the true deployment setting as possible, there are several potential limitations of the present study, providing viable avenues for future work. 

\subsection{Randomizing at the instance level instead of at the user level limited the hypotheses we could evaluate}
While the ideal experiment would not only randomize at the \textit{transaction level}, as we did in our study, it would also include a large number of users to allow for randomization at the \emph{user level}. However, identifying such settings poses practical challenges because, in general, there is limited availability of users in most real-world contexts with sufficient domain expertise who can serve as participants. 
Thus, an important limitation of this study is the small number of analysts (3 analysts) in our experiment.
Though we did not evaluate user-level hypotheses, randomizing at the transaction level still enabled us to test hypotheses related to explainable ML method usefulness in the context of fraud detection. 
We note that our experiment reflects the deployment context as it used \textbf{all the analysts available for the deployment context}. The partnering organization has a staff of internal analysts with expertise in e-commerce fraud, all of whom participated here.
Identifying real-world contexts that make larger-scale studies possible would be a promising avenue for future work to explore potential interactions between the utility of explanation methods and individual characteristics (e.g., whether less experienced users benefit more from these methods). 


\subsection{Short duration of each experimental alleviating opportunities for feedback}
An additional limitation was the relatively short time the analysts spent with each arm and the lack of opportunities for intermediate feedback on their performance. 
In our follow-up interviews, the analysts highlighted one important difference between the experimental setup and real-world deployment, which was that real-world deployment included an ongoing process of learning and improving by receiving periodic feedback both on their overall performance as well as specific examples they classified incorrectly. 
This poses an interesting  avenue for future work: although the ML explanations didn't seem helpful for the analysts here, perhaps they could learn to make better use of this information over time. 
Such an experiment might involve giving the analysts a period of initial training with each arm after which they receive a performance review, where experimental differences are measured only on subsequent decisions. 

\subsection{Focusing only on post-hoc feature-based explanations}
We only focus on post-hoc explanation methods and while the performance of the three methods considered here was similar, it remains possible that other modalities of offering explanations might be more useful for this particular task. 
Future work can explore other explainable ML methods such as inherently-interpretable models \cite{zeng2017riskslim, caruana2015intelligible, lakkaraju2016interpretable}, counterfactual explanations \cite{mothilal2020dice, poyiadzi2020face}, and example-based methods \cite{kim2016examples}. 
It is possible that the fraud detection setting is not conducive to post-hoc feature-based explanations as the review band consists of transactions where the model is less confident. 
This is also an opportunity for additional methodological research, in both the areas of explainable ML and human-computer interaction: can we design explanation methods (and associated user interfaces) that are well suited to helping users with domain expertise make better decisions in the face of model uncertainty?

\section{Summary}
In this work, we took steps toward filling a gap in the field of explainable ML --- designing an experiment for evaluating the real-world efficacy of explainable ML methods that closely reflects the deployment context by making choices about tasks, data, and users that are grounded in the operational objectives of the use-case.
We built on and improved an experimentation setup from a previous study \cite{jesus2021} where fraud analysts make fraud detection decisions with the aid of ML predictions and explanations by modifying the experimental design to better reflect the deployment scenario. 
We believe that this work represents the most robust experimental setup for evaluating post-hoc explainable ML methods reported to date by combining a real task, real-world data, and real users with domain expertise.

Our findings highlighted several points: (1) A need to design methods that tackle specific real-world use-cases, rather than trying to develop ``general purpose'' explainable ML methods that lack grounding in the requirements of practical applications. 
We observed that the tested post-hoc explanation methods did not improve analysts' performance. 
In fact, compared to showing them only the ML score, the explanations made the analysts slower, thus reducing business utility; (2) The critical importance of designing experiments that reflect the deployment context. The modifications we made to the experiment setup resulted in vastly different conclusions to the ones made in the previous study applying the same post-hoc explanation methods to the same fraud detection context; (3) The importance of going beyond confusion matrix-based metrics and choosing metrics that reflect operational objectives. We observed that improvements in the accuracy metrics did not necessarily translate to improvements in the PDR metric that captured business utility; And, (4) the disconnect between objective performance metrics and subjective self-reported measures of explanation utility. Fraud analysts reported high levels of confidence when they were shown explanations compared to the ML model. However, we did not observe any significant differences in their behavior in terms of decision rates, or actual performance in terms of operational metrics.
Beyond the context of fraud detection, we hope that this work 1) provides a template for experimental design and lessons for future evaluations of explainable ML methods in other contexts for explainable ML researchers, and 2) a case study for practitioners exploring the use of explainable ML methods in real-world settings.

\bibliographystyle{ACM-Reference-Format}
\bibliography{references}


\newpage

\appendix

\section{Analyst Interviews} \label{sec:analyst_interviews}
The primary goal of the interview was to better understand the analyst behaviors in the experiments. We treated the interviews as informal discussions to get a better perspective on the analysts' backgrounds and have candid conversation about their experience participating in the experiments. The main topics we covered in these discussions were:
\begin{enumerate}
    \item How the analysts approached the experiment and viewed their role in the experiment
    \item What the analysts believe they are optimizing for
    \item How the analysts believe the explanations impact their decisions
    \item How the analysts would recommend designing new explanations that would be more suitable for their task
\end{enumerate}

\textbf{1. How did the analysts approach our experiments?} The analysts all said that they were playing the role of a client fraud analyst in our experiment. This means that they tried to use the information provided to build a mental picture of the transactions, which involves looking at the customer's details, the history of the card, and so on. However, one analyst mentioned that this experiment was much lower stress than reviewing real transactions because we gave them much fewer transactions to review in the time allotted. 

\textbf{2. What did the analysts believe they were optimizing for?} In general, the analysts tried to make the right decision, which effectively means being as fast and precise as possible. However, the analysts made it clear that, in practice, an analyst would receive frequent guidance from their managers about what their current goal should be. This means that the ``metric'' in the analyst mind is very client-dependent as well as time- and context-dependent. For example, an analyst might specialize in high versus low value transactions or transactions that belong to a certain geographical region, each of which has its own specific traits. Further, depending on the time of year (i.e. Black Friday versus non-holiday times), there might be different criteria for how quickly analysts would need to turn around reviews. While we discussed a set of concrete, quantitative operational metrics in Section \ref{sec: modifications}, it's clear that there are nuances that a single metric alone would not be able to capture.

\textbf{3. How the analysts believe the explanations impact their decisions?} One analyst said that the explanations can be helpful for pointing to parts of the data to look at further. However, whether or not they considered the explanation more carefully would depend on whether the explanations consistently made sense with their prior knowledge and experience. Another analyst mentioned that if the explanation did not make sense, then they would not continue using the explanations and were able to be much faster at reviewing transactions that way. However, a third analyst was careful to read the explanations and double check the explanations, but mentioned that they would get faster reviewing transactions if they gained trust in the explanation over time. 

\textbf{4. What do the analysts believe can be improved about the explanations?} A common theme mentioned by the analysts is that the explanation should be made so that they can easily verify against other parts of the data or they can verify if the reason aligns with their individual or their team's common decision-making practices. Further, if the interface included additional information about how accurate that explanation was for previous transactions, that would help them gain trust over time. Finally, one analyst suggested modifying the interface so that all of the importance decision-making points onto one window, to avoid scrolling, would allow them to make quicker decisions.

\subsection{Other experimental design lessons based on analyst interviews}

A few other minor lessons regarding experimental design seem worth noting here as they may be helpful for researchers and practitioners developing evaluations in other contexts. First, although all three analysts suggested in follow-up interviews that they saw their role in the experiment as reviewing transactions just as they would in a live setting, it does also seem that at least some of them went to an effort to check and confirm the feature values given in the explanations, which they might not have spent the time to do in production (and may explain some of the slower decision times observed with explanations here). Ensuring that the instructions to the analysts clearly remind them that they should treat the experimental examples just like any other transactions and focus on the same KPIs that they normally would may be particularly helpful here. 

\section{Extended Methods}
\label{sec:methods_appendix}
In this section we provide supplemental information about the analyses we conducted in metric calculation, and ensuring the validity of the conclusions. 

\subsection{Operationalizing the Performance Metrics}

Calculating the metrics (PDR and Decision Time) requires selection of values for the parameters $\alpha, \beta, \delta, \lambda $ and $ \tau $. 
We approximated these values by analysing the available transaction data and consulting analysts after the experiment was completed:

\begin{itemize}
    \item $\lambda = 3$: On average, we observed that the customers returned to the merchant 2.3 times and generated revenue in a period of one year. As we are interested in measuring the value a customer generates over several years, we approximate the expected long term value of each customer to be three times the transaction value.
    \item $\alpha = -3$: After consulting the fraud analysts, we learned that a FN (missing fraud) carried more severe consequences than a FP, due to the short- and long-term impacts on the merchant we elaborated above. As such, we assign a penalty of three times the transaction value.
    \item $\beta = 0.5$: In the case of a FP, we expect the customers to use a different credit card and complete the purchase. However, it was not possible to isolate the probability of a FP leading to a sale or missed sale, just by using the available data. Therefore, we treat those two as equally probable outcomes. 
    \item $\delta = 0.1$: 
    After consulting the analysts, we learned that the merchant in question is confident of customer retention, and losing customers due to FPs is not highly likely. Therefore, we use a low probability of losing the long term value of a customer.
    \item $\tau = 600s $: In our discussions with the fraud analysts, we learned that when a transaction is elevated to a senior analyst, given their diligent inquiries (e.g., contacting the user, investigating their online presence), there is a high likelihood that the correct decision is reached. Therefore, we consider escalated transaction as correctly classified with a 10 minute penalty in time to account for the escalation process. 
\end{itemize}

\subsection{Propagation of errors for estimating metric uncertainty}
As we were deriving the PDR metric from the observed variables, in order to capture the uncertainties of the metric value, we propagated the observed variable uncertainties to the calculated metrics. 
For deriving the confidence intervals for PDR, we followed the following process. 

Let total revenue generated by the decisions be $ G $ and total possible revenue be $ P $. Then, our PDR metric ($ M $) is given by: 
\begin{equation}
    M = 1 - \frac{G}{P}
\end{equation}
The variance of the ratio can be estimated as follows \cite{kendal1994advanced}:
\begin{equation}
    Var\left(\frac{G}{P} \right) \approx  \frac{\mu_g^2}{\mu_p^2} \left[ 
        \frac{\sigma_g^2}{\mu_g^2} + \frac{\sigma^2_p}{\mu_p^2} - 2\frac{cov(G, P)}{\mu_g\mu_p}
    \right]
    \label{eq: var_ratio}
\end{equation}
where $ \mu_g $, $ \mu_p $, $ \sigma^2_g $ and $ \sigma^2_p $ are transaction level means and variances of the generated and the possible revenues in the dataset. 
Furthermore, It is worth noting we constructed confidence intervals using bootstrapped samples (1000 iterations with a sample size of 500) and ended up with a very similar variance estimate for the PDR metric as calculated using equation \ref{eq: var_ratio}.

\subsection{Statistical tests}
\label{sec:stattests_appendix}
In testing our hypotheses, we used several statistical tests depending on the metric of interest. For testing continuous metrics (PDR \& Decision Time), we used \textit{Student's t-test} \cite{student1908probable} for pairwise comparisons between groups. 
For instance-wise binary metrics such as confusion matrix-based metrics (e.g., accuracy, FPR, Recall) and decision rates, we used \textit{Chi-square tests} \cite{pearson1900X} for pairwise comparisons between groups. For all the tests, we used a significance level of 0.10. 
In addition to using statistical tests for group comparisons, we conducted analyses with generalized linear models to account for the variance in analyst performance. As mentioned in the main text, we were limited to three professional fraud analysts; to ensure that the results were not due to aberrations in analyst performance, we used linear models to control for analyst contributions. We used multiple linear regression models to test if the treatment (e.g., showing the ML score and showing the explanations) impacted the decision time while controlling for the analyst and the transaction value. Even when accounting for the variance across analysts, the linear regression model confirmed our findings presented in the main text: Showing the ML model score significantly reduced the decision time ($\beta = -27.9$, $p=.001$), showing explanations from TreeSHAP ($\beta = 15.85$, $p=.02$) and TreeInterpreter ($\beta = 14.24 $, $p=.03$) slowed down the analysts. We used Logistic Regression models for testing the impact on the confusion matrix-based metrics, as with decision-time, our findings were confirmed even when accounting for the variance across analysts: Showing the ML score improved analyst accuracy ($ or = 1.4$, $p=.03$), analysts rejected fewer ($or=0.7$, $p=.03$) and approved more ($or=1.67$, $p=.001$) transactions with the ML model scores compared to data.  

\begin{figure*}
   \centering
   \subfigure[]{\includegraphics[width=0.48\linewidth]{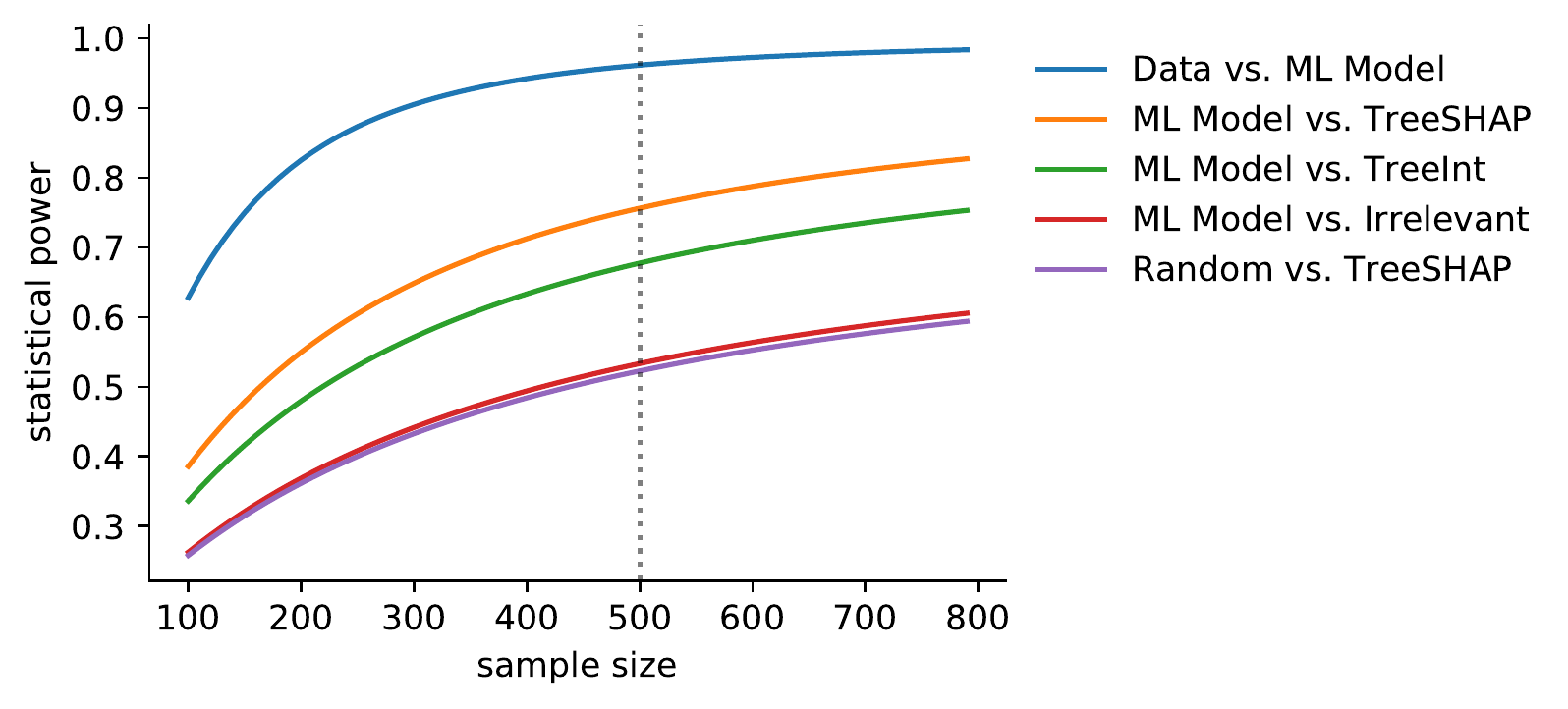}}
   \hfill
   \subfigure[]{\includegraphics[width=0.48\linewidth]{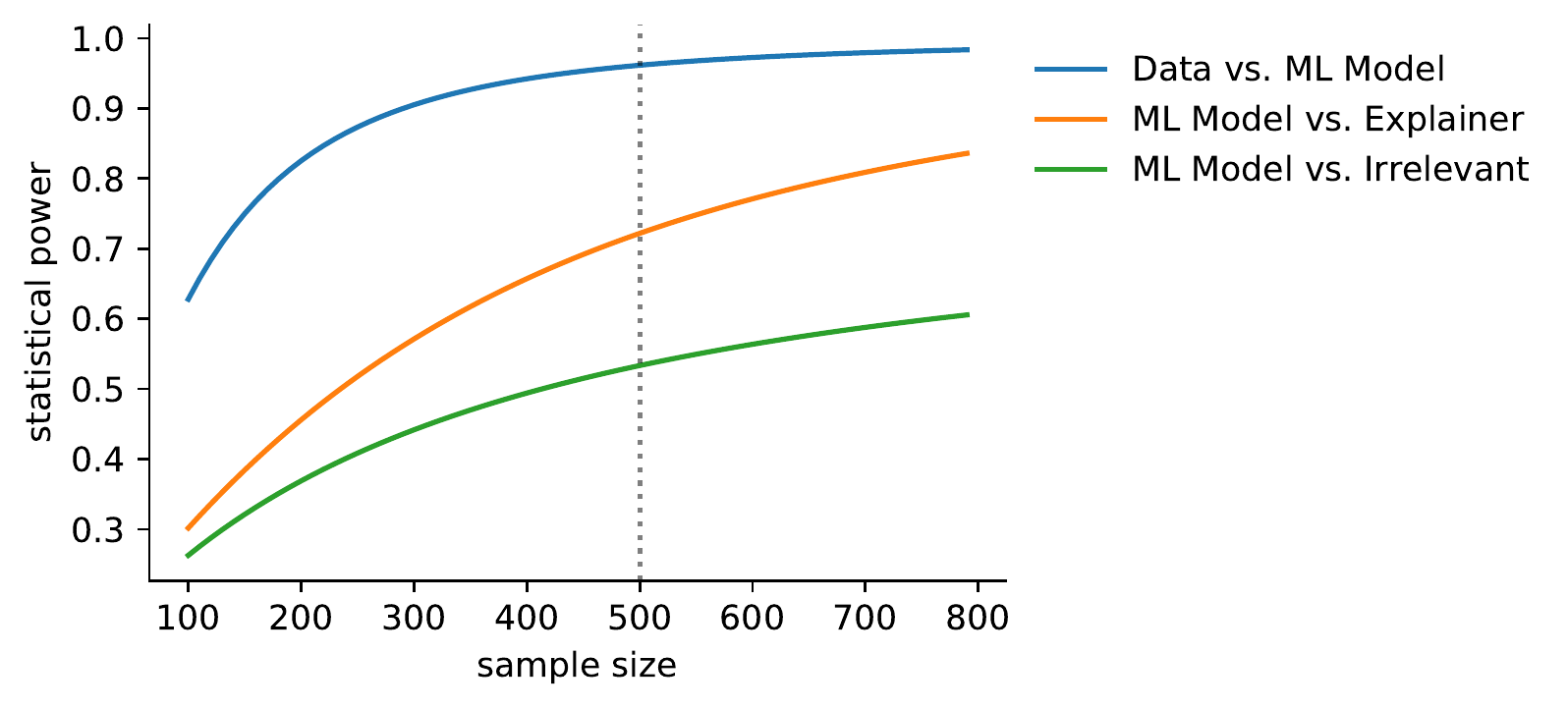}}
    \caption{Statistical Power curves for two group independent sample t-test for the decision time metric (a) when comparing all experiment arms and (b) when all explanation arms grouped into one. We only plot curves for the comparisons that yielded statistically significant differences. The dotted line shows the used sample size. 
    } 
    \label{fig:statpower_time}
\end{figure*}

As noted in the main text, we increased the sample sizes in the presented work compared to the previous study done by Jesus et al.\cite{jesus2021}, with 500 samples per arm (compared to 200 for control arms \& 300 for explainer arms used in the previous study). Figure \ref{fig:statpower_time} shows the statistical power curves for the Two-group independent sample t-tests we conducted to pairwise decision-time comparison. We can see that the effect we observed for the \textit{ML model} over \textit{Data} has high power (90\%), while the effect of \textit{Explanation} arms versus \textit{ML Model} has a power of almost 70\%. Therefore we can be sufficiently confident about the conclusions we have made with respect to our hypotheses. 
We can see that the differences we observed between \textit{Random} \& \textit{TreeSHAP} and \textit{ML Model} \& \textit{Irrelevant} are under powered compared to the other tests with significant differences (~50\%), which is inline with our conclusions as we did not observe strong effects when we compared any of the Stage 3 arms across each other or to Stage 2 (ML Model). Please note that we only report the power for decision time as we did not observe any statistically significant difference for the PDR metric across any arms. 

\subsection{Testing PDR's sensitivity to the parameters}
\label{sec:pdr_sensitivity_appendix}
We determined the values for the PDR parameters based on historical data and conversations with the analysts and other stakeholders. However, because there is some uncertainty in these values, we wanted to understand the sensitivity of our results to their exact specification. To do so, we conducted a parameter grid-search across a range of reasonable values for each parameter. We focused especially on the significance of differences in PDR values between the experiment arms. 

\begin{figure*}
   \centering
    \includegraphics[width=\linewidth]{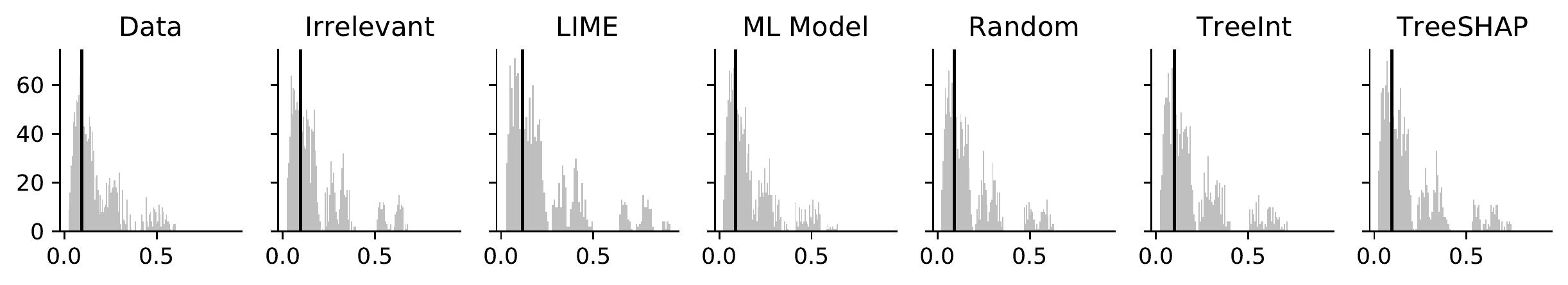}
    \caption{The distributions of the PDR metric resulted in the sweep of the parameters for each experiment arm. The vertical lines show where the PDR value resulted by our current choice of parameters fall in the distribution. }
    \label{fig:pdr_dist_ps}
\end{figure*}

See Table \ref{tab:paramsweep} for the range of different values for each of the parameters considered. We did not observe any statistically significant differences in PDR for any of the parameter combinations we tested. Figure \ref{fig:pdr_dist_ps} shows the distribution of PDR values resulted from all different combinations of parameter values. Further, the figure shows where the current parameter values place with respect to the distribution. 
We can see that PDR current values fall within the dense area of the distribution.

\begin{table*}[]
    \centering
    \caption{The parameter value ranges that we used in performing the sensitivity analysis of PDR. The values that were used for the main calculations are given in bold}
    \begin{tabular}{c l  l}
    \toprule
        Parameter & Description & Explored Values \\
    \midrule
        $ \alpha $ & Weight parameter for FNs & -1, -2,  \textbf{-3}, -5, -6 \\
        $ \beta $  & Prob. of losing the transaction given an FP & 0.3, 0.4, \textbf{0.5}, 0.6 \\
        $ \delta $ & Prob. of losing the customer given an FP & 0, 0.05, \textbf{0.1}, 0.2 \\
        $ \lambda $ & Long term customer worth & 0, 1, \textbf{3}, 5\\
        $ \tau $ & Time penalty for transactions marked as suspicious & 0, \textbf{600}, 1800 \\
    \bottomrule
    
    \end{tabular}
    \label{tab:paramsweep}
\end{table*}

\subsection{List of Irrelevant Features}
\label{sec:irrelevant-appendix}

\begin{table}[]
\caption{Examples of features created to generate irrelevant explanations. All the features were verifiable from the data, but it was ensured that they were not correlated with the outcome}
\label{tab:irrelevant_features}
\begin{tabular}{l | l}
\toprule
Data Field                 & Feature                                             \\
\midrule
Transaction Timestamp      & Day of the week                                     \\
Transaction Timestamp      & Month                                               \\
Transaction Timestamp      & Day of the month                                    \\
Transaction Timestamp      & Hour (0-60)                                         \\
Transaction Timestamp      & Minute of hour (0-60)                               \\
Transaction Timestamp      & Second (0-60)                                       \\
Payment Method             & Last two digits of card                             \\
Payment Method             & Month of card expiry                                \\
Name                       & Middle name                                         \\
Name                       & Number of names                                     \\
Addresses                  & House number                                        \\
Addresses                  & Street Suffix                                       \\
IP address                 & Last two digits of IP address                       \\
Emails                     & First character of email                            \\
Emails                     & Length of email prefix length                       \\
Emails                     & Whether the first name appears in the email address \\
Transaction details        & Number of cents in the transaction amount           \\
Transaction details        & Whether the transaction amount contains a 4         \\
Transaction details        & Last two digits of the transaction ID               \\
Transaction details        & Item name contains word `X'                         \\
Transaction details        & Transactions contain k items n-gram `X'             \\
Transaction details        & Last two characters of the item name                \\
User account creation date & Day of the week                                     \\
User account creation date & Month                                               \\
User account creation date & Day of month                                        \\
User account creation date & Hour                                                \\
User account creation date & Minute of hour (0-60)                               \\
User account creation date & Second (0-60)                                       \\        
\bottomrule
\end{tabular}
\end{table}

Table \ref{tab:irrelevant_features} lists the set of irrelevant features that were used to generate explanations. For each transaction, 6 random features from this set was sampled and each feature was assigned a random feature importance polarity.


\section{Additional figures}

\begin{figure*}
   \centering
    \includegraphics[width=\textwidth]{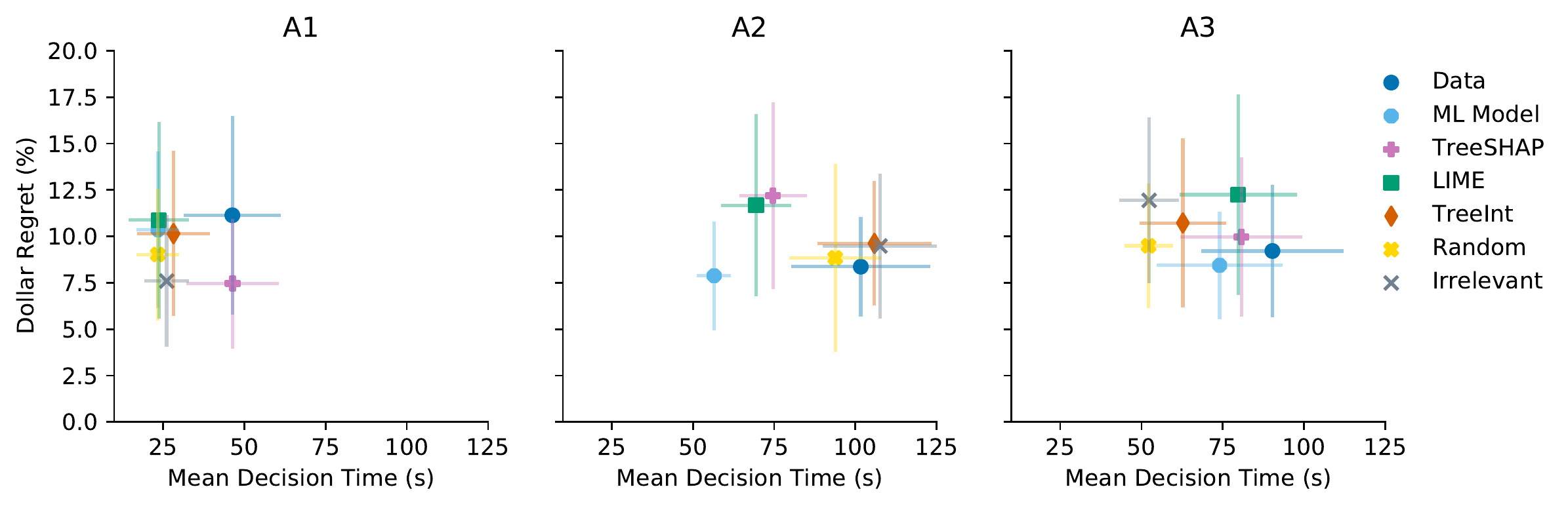}
    \caption{PDR vs Mean Decision Time by individual analyst. We can see the variance across the three users (A1, A2, and A3), especially in time taken to make decisions. However, we did not see significant differences across the analysts with respect to the PDR metric}
    \label{fig:pdrtime_users}
\end{figure*}

\begin{figure*}
   \centering
   \subfigure[]{\includegraphics[width=0.48\linewidth]{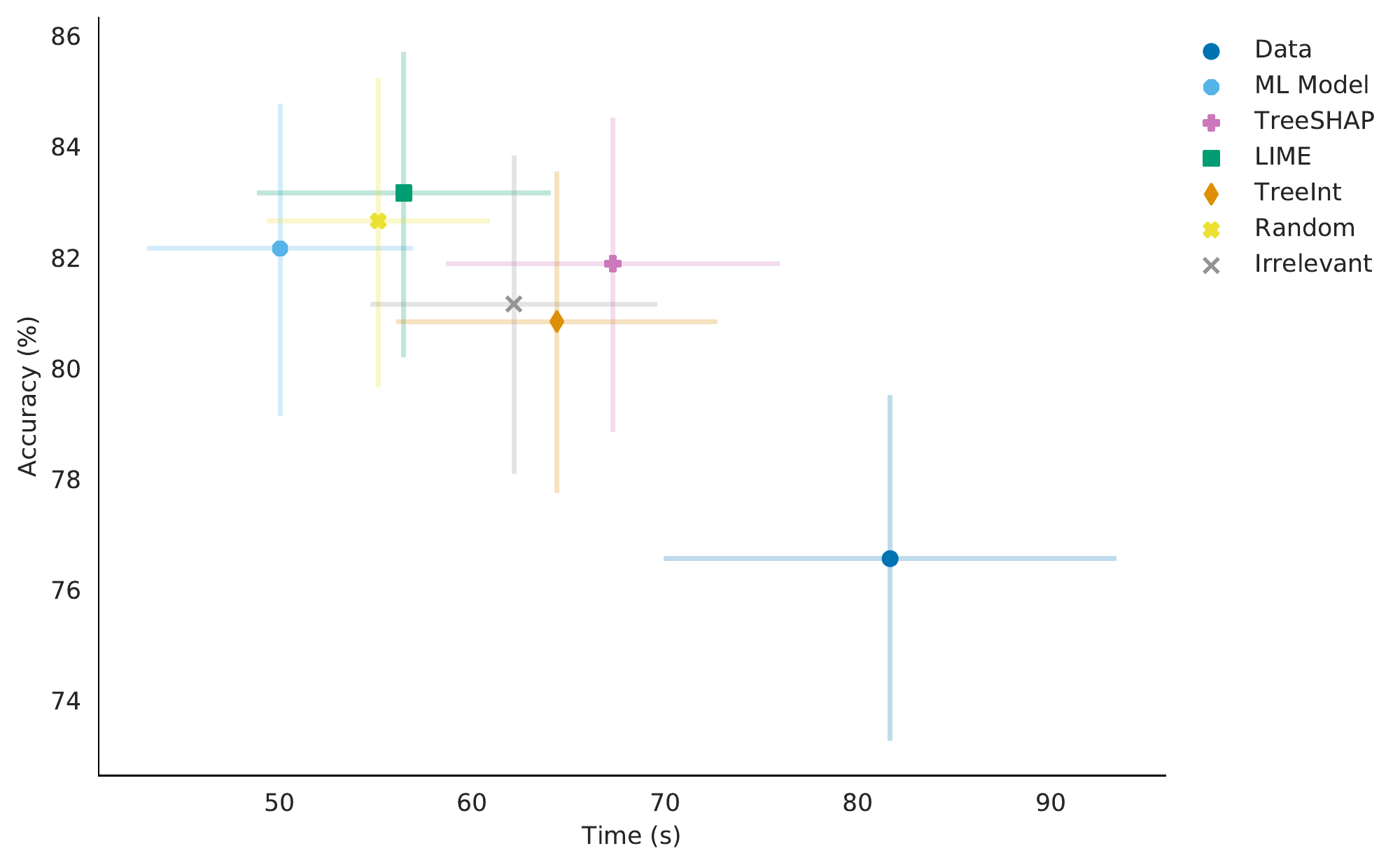}}
   \hfill
   \subfigure[]{\includegraphics[width=0.48\linewidth]{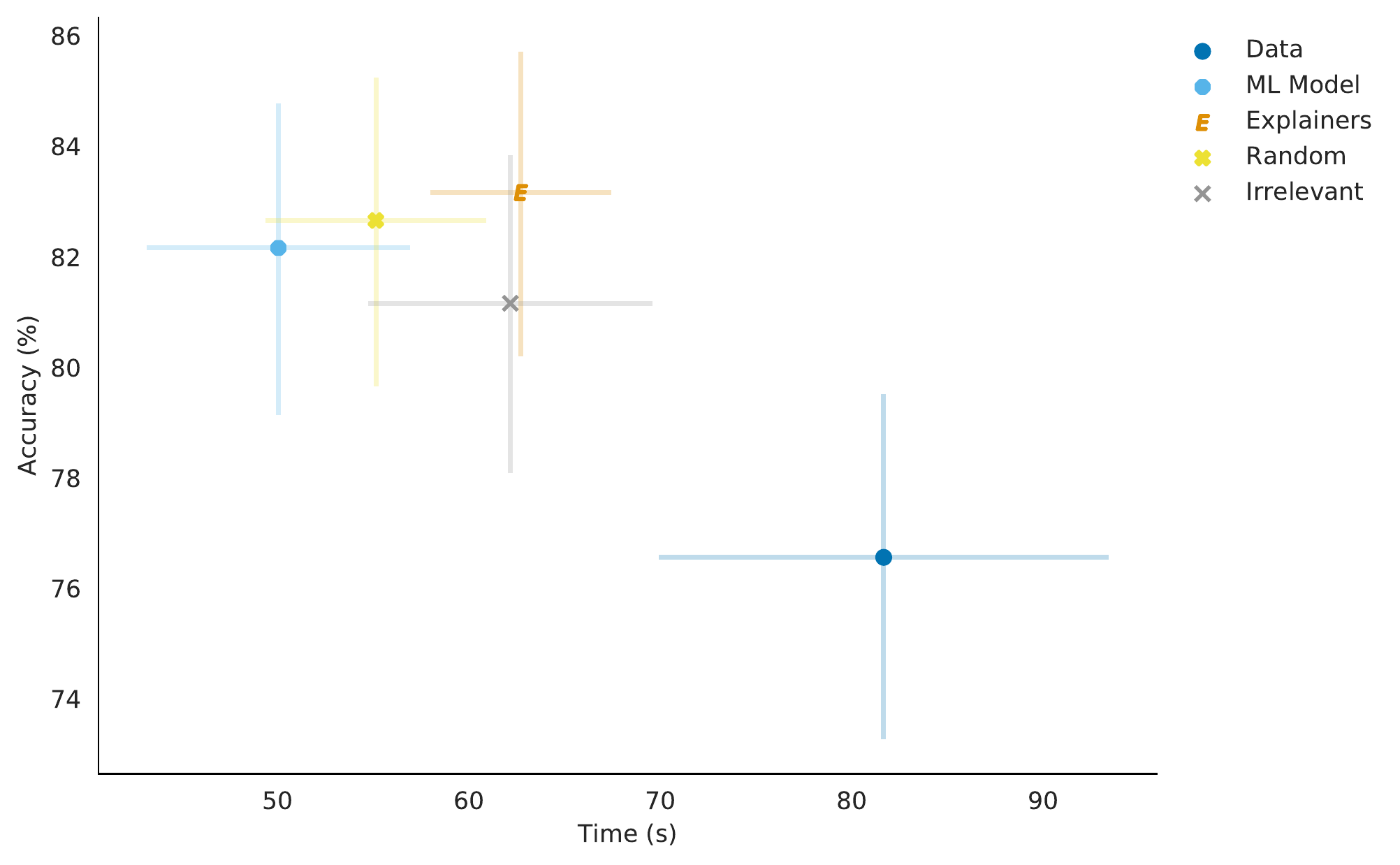}}
    \caption{ Decision Accuracy versus Decision Time  (a) for all experiment arms, (b) with all explanation arms grouped into one. Error bars show the 90\% confidence intervals.  It is worth noting that accuracy is not well suited in this context due to the skewed label distribution.
    } 
    \label{fig:acc_time}
\end{figure*}

\begin{figure*}
   \centering
   \subfigure[]{\includegraphics[width=0.48\linewidth]{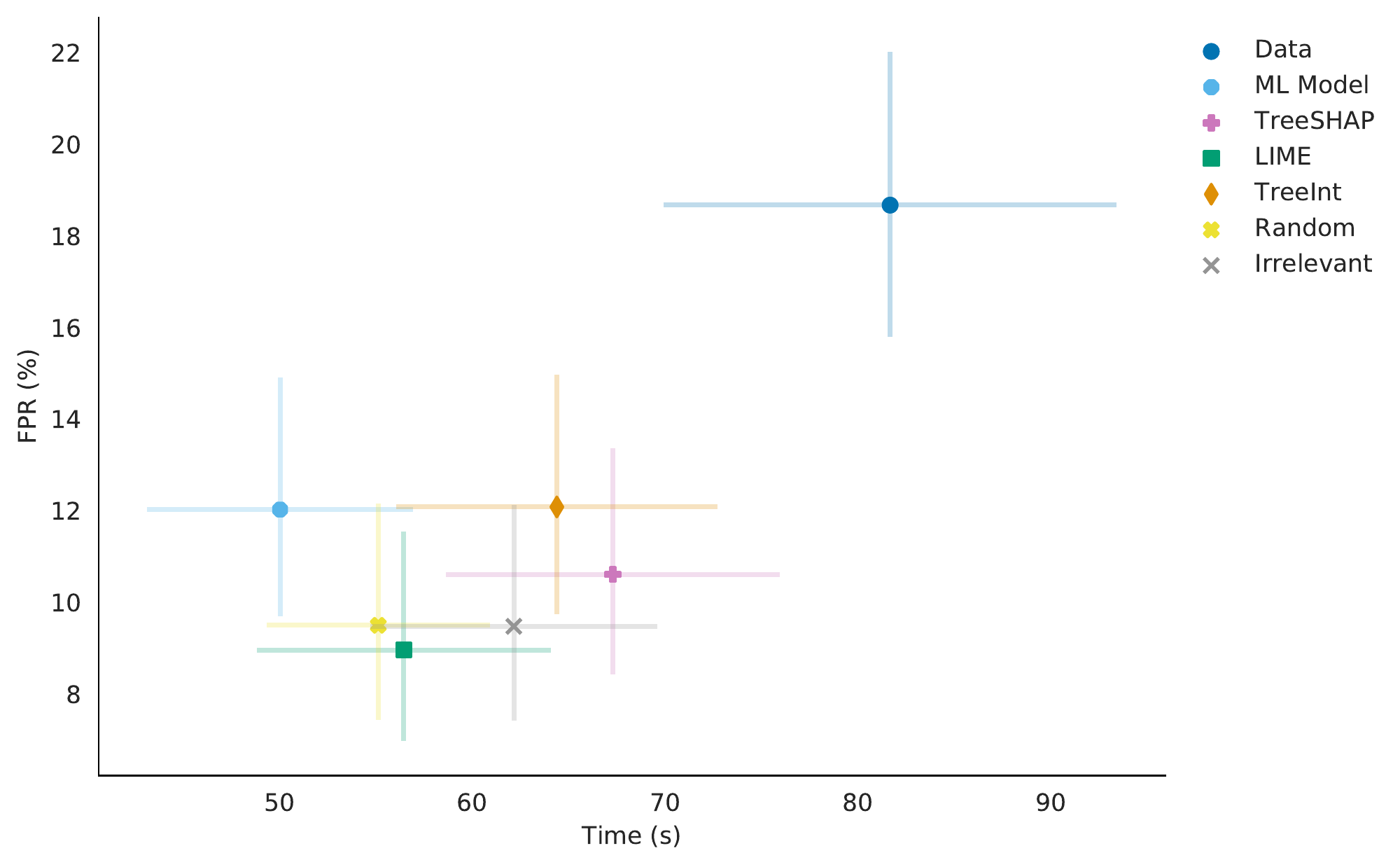}}
   \hfill
   \subfigure[]{\includegraphics[width=0.48\linewidth]{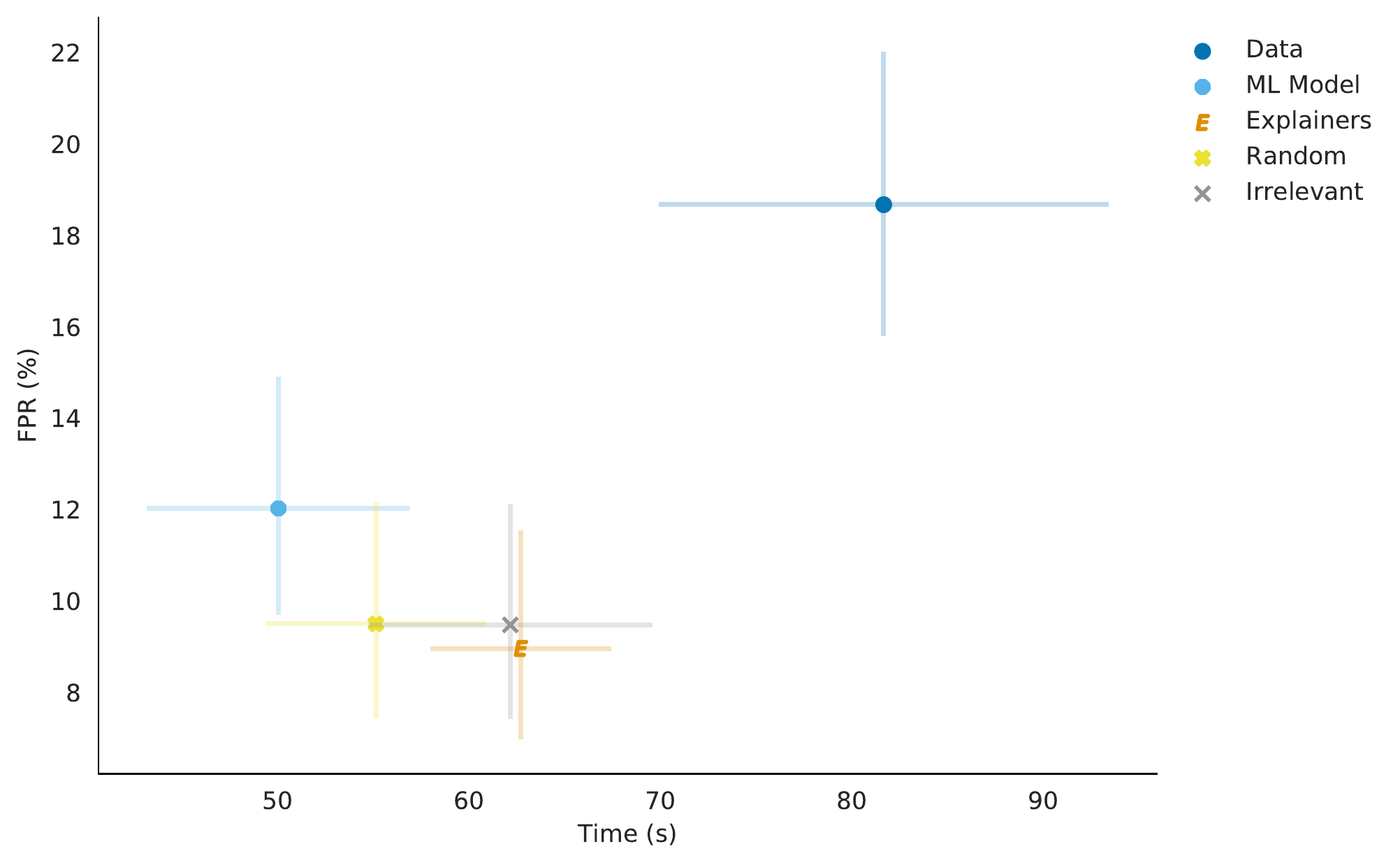}}
    \caption{ False Positive Rate versus Decision Time  (a) for all experiment arms, (b) with all explanation arms grouped into one. Error bars show the 90\% confidence intervals.
    } 
    \label{fig:fpr_time}
\end{figure*}

\begin{figure*}
   \centering
   \subfigure[]{\includegraphics[width=0.48\linewidth]{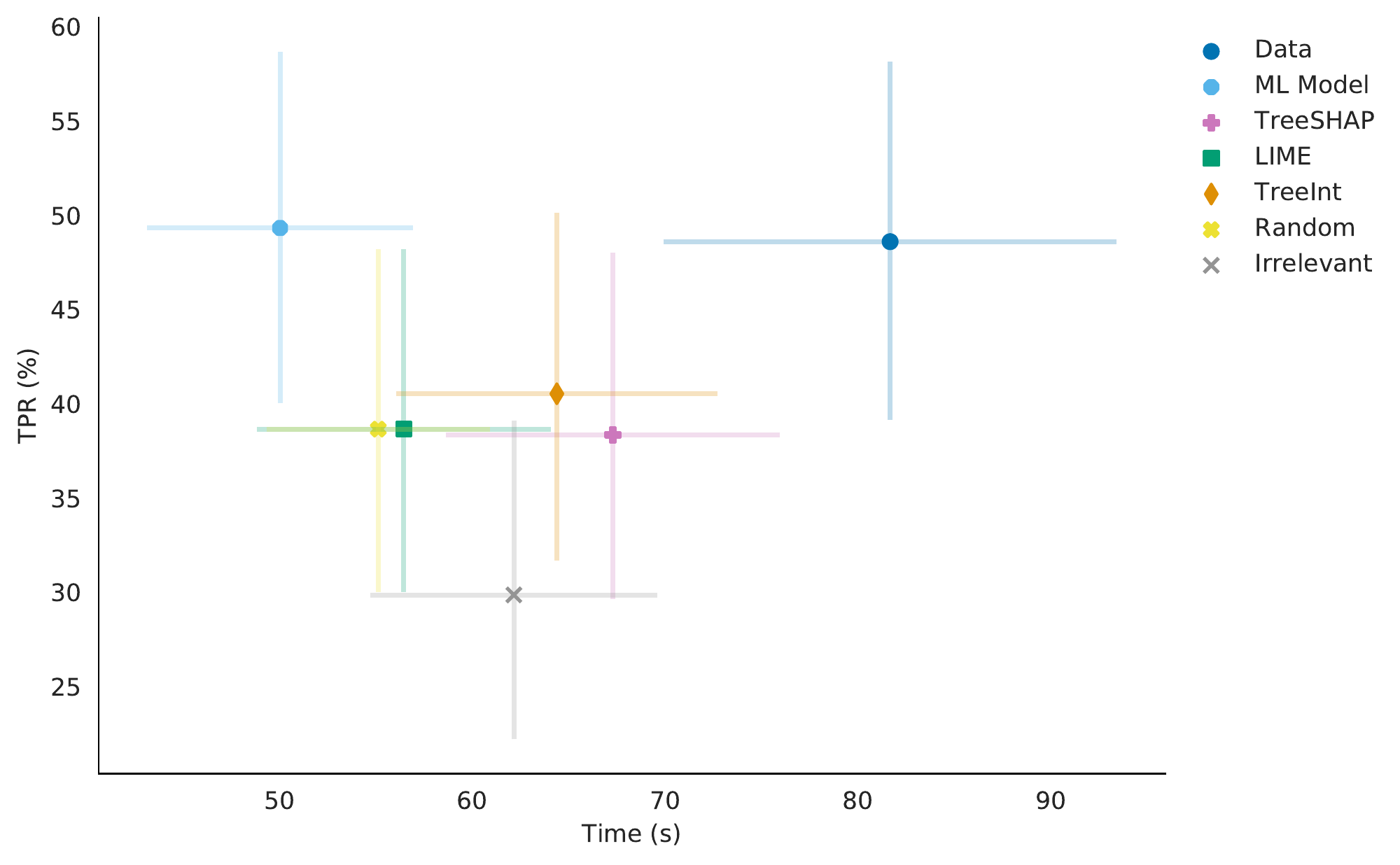}}
   \hfill
   \subfigure[]{\includegraphics[width=0.48\linewidth]{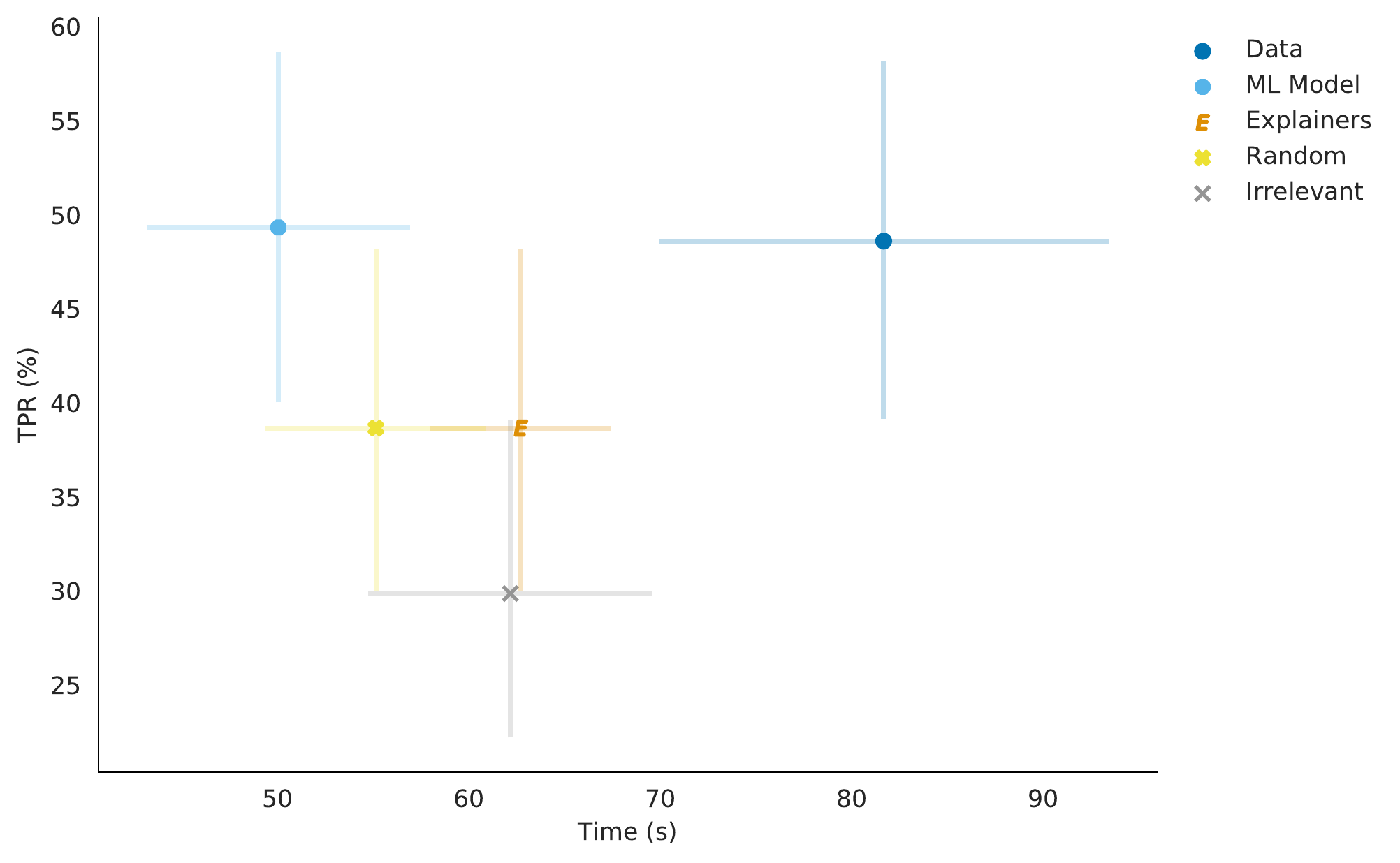}}
    \caption{ True Positive Rate versus Decision Time  (a) for all experiment arms, (b) with all explanation arms grouped into one. Error bars show the 90\% confidence intervals. We can observe the similarity in the TPR values between the \textit{Data} and \textit{ML Model} arms contributing to the similar PDR values that weighs the FN rate over the FP rate.  
    } 
    \label{fig:acc_time}
\end{figure*}

\begin{figure*}
   \centering
    \includegraphics[page=1,width=\linewidth]{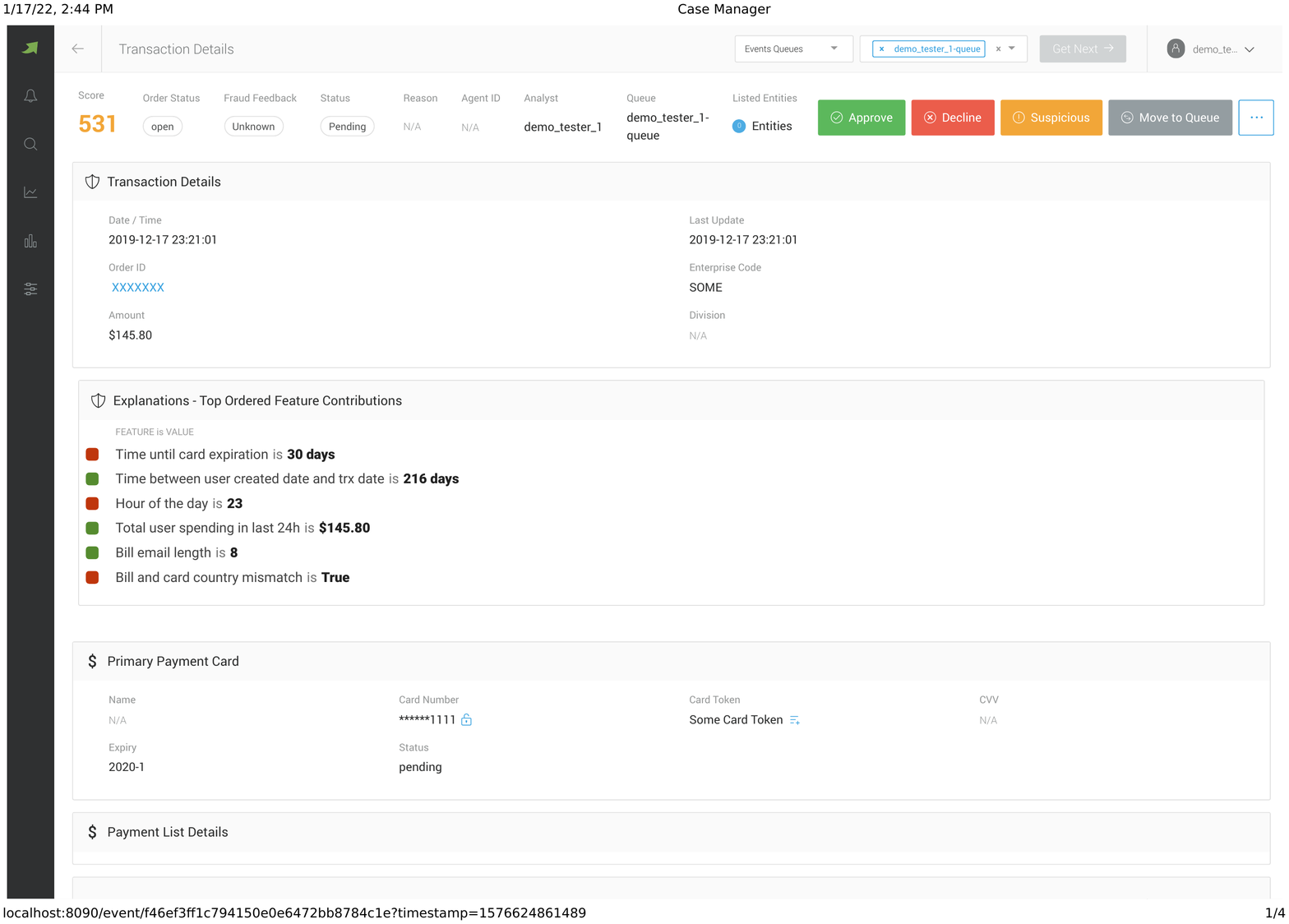}
    \caption{Example of the full interface used by fraud analysts, populated with sample data.}
    \label{fig:full_interface}
\end{figure*}

\begin{figure*}\ContinuedFloat
   \centering
    \includegraphics[page=2,width=\linewidth]{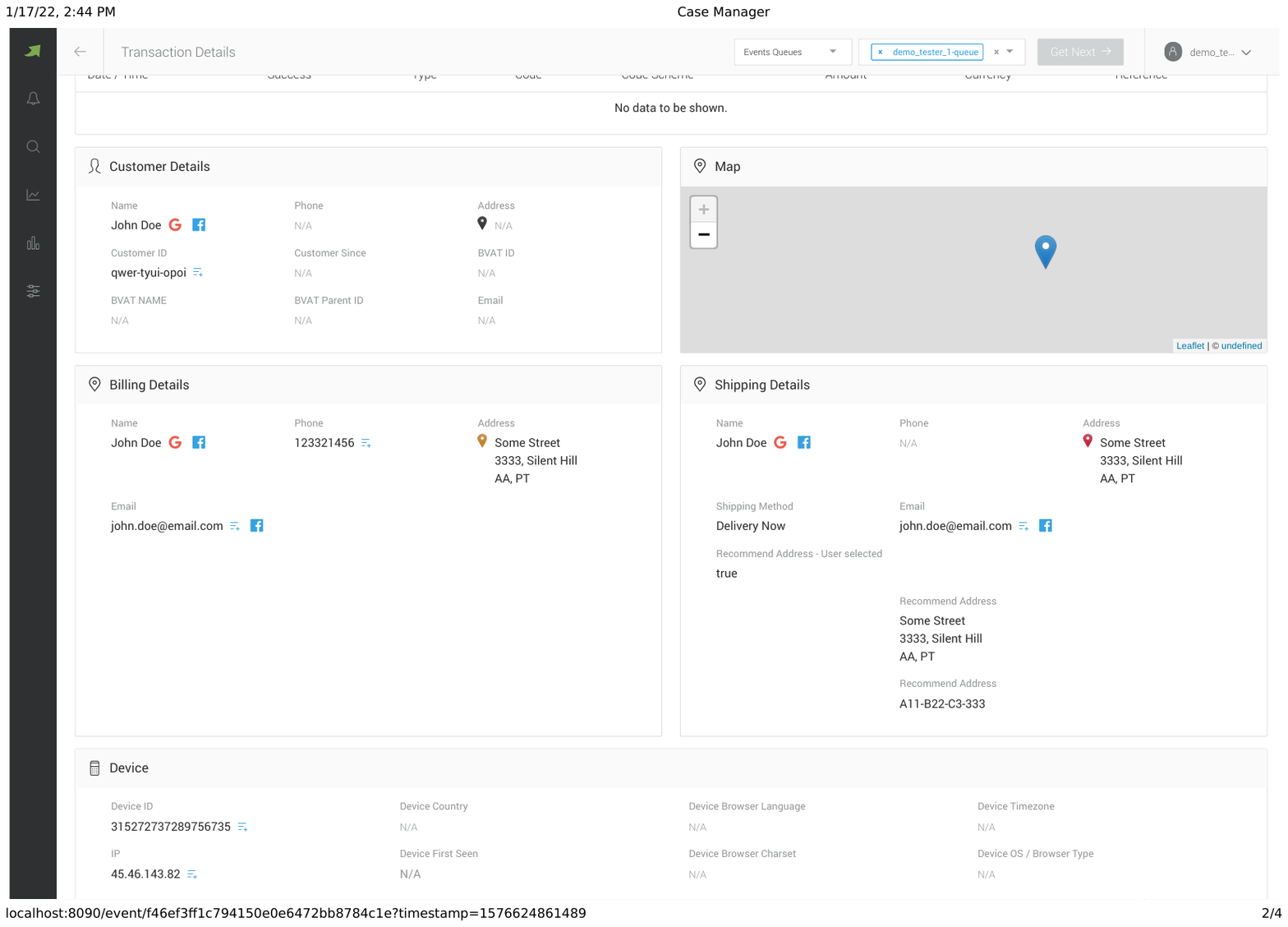}
    \caption{Example of the full interface used by fraud analysts, populated with sample data (continued).}
\end{figure*}

\begin{figure*}\ContinuedFloat
   \centering
    \includegraphics[page=3,width=\linewidth]{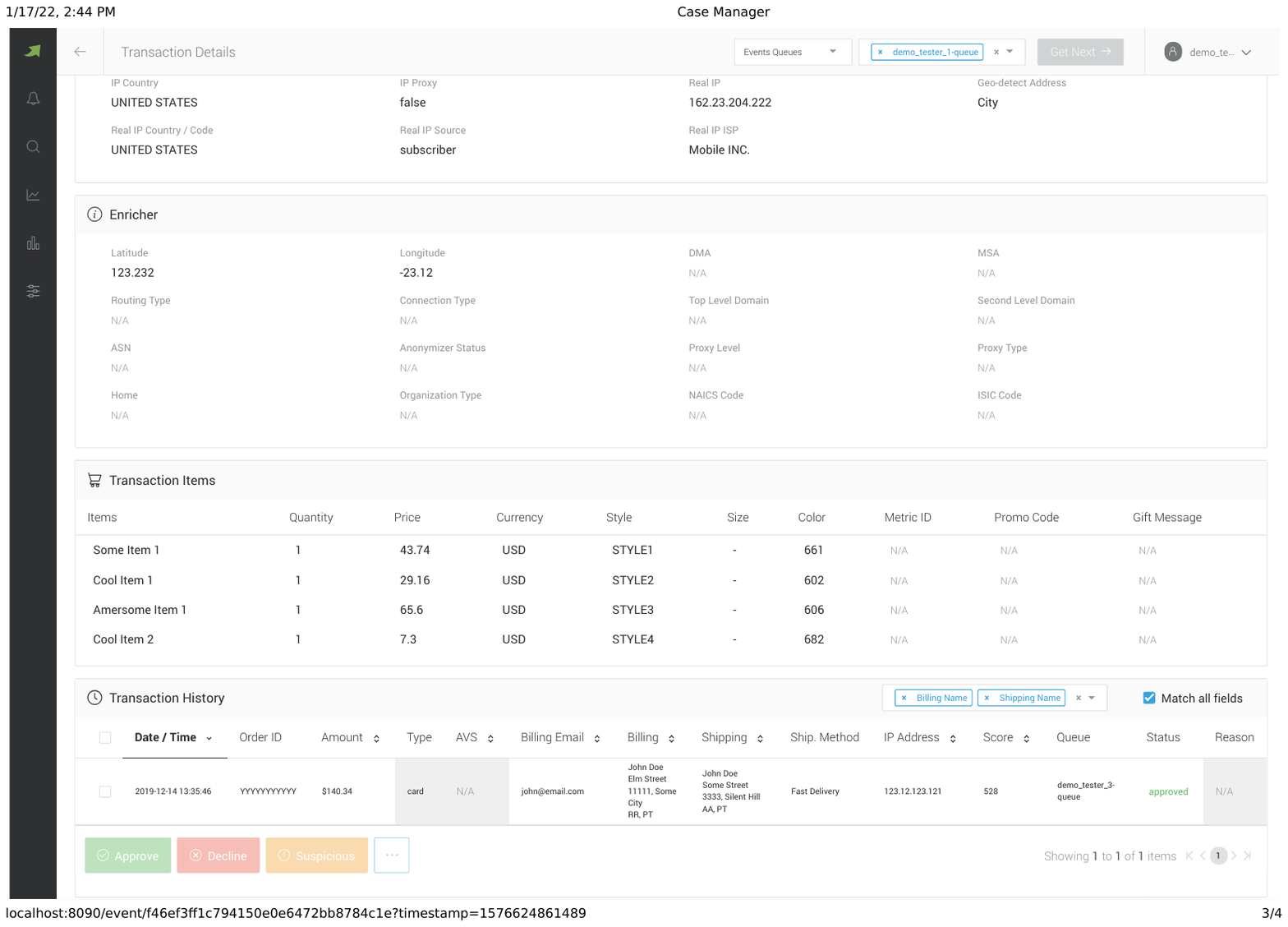}
    \caption{Example of the full interface used by fraud analysts, populated with sample data (continued).}
\end{figure*}

\end{document}